\RequirePackage{tikz}
\documentclass[sn-mathphys]{sn-jnl}

\usepackage{tikz}
\usepackage{pgfplots}
\pgfplotsset{
    every axis plot post/.style={
        line join=round
    }
}
\usepackage{scalefnt}
\usepackage{subcaption}
\usepackage{bbm}
\usepackage{mathtools}
\usetikzlibrary{matrix}
\usetikzlibrary{backgrounds}
\usetikzlibrary{calc}
\usetikzlibrary{shapes.misc}
\usetikzlibrary{shapes}
\usetikzlibrary{decorations.pathmorphing}
\usetikzlibrary{patterns}

\jyear{2023}%

\theoremstyle{thmstyleone}%
\theoremstyle{thmstyletwo}%

\theoremstyle{thmstylethree}%
\newcommand{\bs}[1]{\boldsymbol{#1}}
\newcommand{\bno}[2]{\bs{#1}_{#2}}
\newcommand{\btx}[2]{\bs{#1}_{\mathrm{#2}}}

\newcommand{\qno}[1]{\bno{q}{#1}}
\newcommand{\qtx}[1]{\btx{q}{#1}}
\newcommand{\xno}[1]{\bno{x}{#1}}
\newcommand{\xtx}[1]{\btx{x}{#1}}
\newcommand{\vno}[1]{\bno{v}{#1}}
\newcommand{\lano}[1]{\bno{\lambda}{#1}}
\newcommand{\hT}{^\mathsf{T}}
\newcommand{\hC}{^\mathsf{C}}

\newcommand{\wno}[1]{\bno{\omega}{#1}}

\newcommand{\Lmat}[1]{\bs{L}(#1)}
\newcommand{\LTmat}[1]{\Lmat{#1}\hT}
\newcommand{\Rmat}[1]{\bs{R}(#1)}
\newcommand{\RTmat}[1]{\Rmat{#1}\hT}

\newcommand{\dt}{\Delta t}

\newcommand{\mrml}[1]{_{\mathrm{#1}}}
\newcommand{\mrmu}[1]{^{\mathrm{#1}}}

\newcommand{\mrmpreu}[2]{\prescript{\mathrm{#1}}{}{#2}}

\definecolor{mycolor1}{rgb}{0.00000,0.44700,0.74100}%
\definecolor{mycolor2}{rgb}{0.85000,0.32500,0.09800}

\newcommand{\myupdate}[1]{#1}

\begin{document}

\title[Maximal-Coordinate Dynamics]{Variational Integrators and Graph-Based Solvers for Multibody Dynamics in Maximal Coordinates}


\author*[1]{\fnm{Jan} \sur{Br\"udigam}}\email{jan.bruedigam@tum.de}

\author[1]{\fnm{Stefan} \sur{Sosnowski}}\email{sosnowski@tum.de}

\author[2]{\fnm{Zachary} \sur{Manchester}}\email{zacm@cmu.edu}

\author[1]{\fnm{Sandra} \sur{Hirche}}\email{hirche@tum.de}

\affil[1]{\orgdiv{School of Computation, Information and Technology}, \orgname{Technical University of Munich}, \orgaddress{\street{Barer Str. 21}, \city{Munich}, \postcode{80333}, \country{Germany}}}

\affil[2]{\orgdiv{The Robotics Institute}, \orgname{Carnegie Mellon University}, \orgaddress{\street{5000 Forbes Avenue}, \city{Pittsburgh}, \postcode{15213}, \state{PA}, \country{USA}}}


\abstract{Multibody dynamics simulators are an important tool in many fields, including learning and control for robotics. However, many existing dynamics simulators suffer from inaccuracies when dealing with constrained mechanical systems due to unsuitable integrators \myupdate{with bad energy behavior and problematic constraint violations, for example for contact interactions.} Variational integrators are numerical discretization methods that can reduce physical inaccuracies when simulating mechanical systems, and formulating the dynamics in maximal coordinates allows for easy and numerically robust incorporation of constraints such as kinematic loops or contacts. Therefore, this article derives a variational integrator for mechanical systems with equality and inequality constraints in maximal coordinates. Additionally, efficient graph-based sparsity-exploiting algorithms for solving the integrator are provided and implemented as an open-source simulator. The evaluation of the simulator shows improved physical accuracy due to the variational integrator and the advantages of the sparse solvers. \myupdate{Comparisons to minimal-coordinate algorithms show improved numerical robustness} and application examples of a walking robot and an exoskeleton with explicit constraints demonstrate the necessity and capabilities of maximal coordinates.}




\maketitle

\section{Introduction}
\myupdate{Simulators for mechanical systems are widely used, for example in testing and verification \cite{agarwal_simulation_2010,kuhn_dynamics_2018}, model-based control strategies \cite{koenemann_whole_2015, erez_integrated_2013}, or learning-based methods \cite{andrychowicz_learning_2020,lee_learning_2020}.
However, many common simulators have numerical difficulties with more complex mechanical systems involving constraints \cite{gonzalez_benchmarking_2006}. Such constraints can represent joints connecting rigid bodies, which may form kinematic loops, for example, in exoskeletons. Constraints can also be used to confine the movement of bodies, for example, to model joint limits in robotic arms, or to describe contact with other bodies or the environment in walking and grasping. Exactly enforcing such constraints can cause numerical issues, for example, due to the stiff nature of contact interactions. To alleviate these numerical issues, simulators often allow small constraint violations by representing all constraints as spring-damper elements as in MuJoCo \cite{todorov_MuJoCo_2012} and Brax \cite{freeman_brax_2021}, or by accepting interpenetration of bodies as in Drake \cite{tedrake_drake_2019} and Bullet \cite{coumans_bullet_2016}. Small violations can sometimes be acceptable, for example, contact interpenetration in the order of micrometers for meter-scale walking robots. But millimeter or centimeter violations, for example in MuJoCo, can be considered too large. Employing these methods and accepting larger constraint violations for stable simulations contributes to the sim-to-real gap, a major issue in robotics \cite{zhao_sim_2020}.}

\myupdate{Therefore, this article addresses physically accurate simulations of mechanical systems. Specifically, we focus on good modeling of the energy behavior of such systems, as this quantity is important for the stability of dynamical systems from a control theory perspective. Moreover, we treat the correct enforcement of constraints since allowing for constraint violation can lead to problematic results. For example, allowing softness in contact interactions can lead to wrong contact points, which may make a robot fail to properly grasp an object when the simulation results are directly transferred to a real-world application.}

\myupdate{Most simulators, including the ones listed above except Brax, use minimal coordinates (generalized coordinates) as a mechanism's state representation. Given a system with $m$ degrees of freedom and a set of constraints that removes $c$ degrees of freedom, the system is parameterized by $n=m-c$ independent coordinates. In minimal coordinates, the constraints are implicitly incorporated into the dynamics equations, and only the smallest possible number of variables is retained, which leads to dimensionally small problem sizes. However, eliminating constraints requires specialized treatment of each constraint type and is not possible for nonholonomic constraints such as contacts. Furthermore, the algorithms typically used for minimal coordinates can have numerical issues such as ill-conditioning \cite{featherstone_rigid_2008, featherstone_empirical_2004}, which may be increased by their recursive nature, although a rigorous study of the numerical effects and stability of minimal-coordinate dynamics algorithms remains to be done. We provide some empirical evidence of these numerical issues in Sec. \ref{sec:evaluation}.}

\myupdate{It is also possible to explicitly retain all or part of the constraints on a mechanical system by introducing constraint forces to adhere to these constraints. With this approach, redundant coordinates are obtained since the constraints make part of the coordinates mathematically dependent. There are many ways to parameterize mechanical systems in redundant coordinates. Natural coordinates use three points in Cartesian coordinates to parameterize the configuration of each body \cite{jalon_twenty_2007}. Alternatively, the position of the center of mass of a body and a director frame for the orientation can be used \cite{betsch_constrained_2001,kinon_ggl_2023}. These methods have in common that they aim at circumventing the inherent intricacies of parameterizing rotations. In contrast, we use so-called maximal coordinates \cite{baraff_linear-time_1996}, which represent each body in a mechanism with its three degrees of freedom for the position of the center of mass and three degrees of freedom for the orientation by purposefully building on the group structure of rotations. All kinematic relations, such as joints or contacts, are formulated as explicit constraints. The use of maximal coordinates allows for very modular modeling of dynamics, which makes it possible to describe many different kinds of constraints in a unified and non-specialized manner. Additionally, resorting to maximal and redundant coordinates enables the use of different algorithms than for minimal coordinates, specifically direct matrix methods, which generally have well-investigated numerical properties and enable the formulation of stable and well-conditioned algorithms \cite{higham_accuracy_2002}. The core idea for direct matrix methods is to modify the underlying matrix before, during, and after it is used in an algorithm to improve the stability and accuracy of an algorithm. Examples of such modifications are scaling certain matrix entries or iterative solution refinement. The few existing works on simulation, specifically in maximal coordinates, have explored algorithms for continuous-time forward dynamics \cite{baraff_linear-time_1996} and discrete-time algorithms for mechanisms without kinematic loops or contacts \cite{brudigam_linear-time_2020}. Besides their application in simulation, maximal coordinates have shown promising results in control applications \cite{brudigam_linear-quadratic_2021,shield_minor_2022}. The simulator Brax is also based on maximal coordinates but suffers from the aforementioned issue of soft constraint handling.}

\myupdate{Besides the numerical issues related to constraints, all of the packaged simulators above use integrators for the dynamics that are not well suited for mechanical systems due to the unrealistic energy behavior of certain explicit or implicit Runge-Kutta methods. Deriving a perfect integrator is generally not trivial \cite{ge_lie-poisson_1988,chartier_algebraic_2006}, but certain classes of integrators exhibit excellent properties, among them symmetric and symplectic methods \cite{hairer_geometric_2006}. An elegant way to derive symplectic integrators for mechanical systems is through a variational perspective, leading to variational integrators. These integrators are derived by discretizing the derivation of the mechanism's differential equations instead of the differential equations themselves. Thus, certain properties such as energy and momentum conservation are maintained \cite{marsden_discrete_2001}. Compared to classical discretizations such as explicit Runge-Kutta methods, larger time steps can be taken due to the increased physical accuracy, and constraint drift is avoided entirely. Computationally efficient variational integrators have been theoretically investigated in minimal coordinates \cite{johnson_scalable_2009,lee_linear-time_2016,fan_efficient_2018}. However, the implementations are limited to simple mechanical systems with few joint types and no contact interactions, potentially due to numerical difficulties. Variational and other structure-preserving integrators have also been investigated in redundant coordinates. Here, much effort has been put into variational integrators for systems with explicit constraints \cite{marsden_discrete_2001,leyendecker_variational_2008,kinon_ggl_2023}. Another investigated approach is nullspace methods which eliminate constraints while remaining structure-preserving \cite{betsch_discrete1_2005,betsch_discrete2_2006}. Redundant coordinates are also well suited for structure-preserving integration of flexible multibody systems \cite{betsch_dae_2002,leyendecker_discrete3_2008,brugnoli_port_2021}. These presented works deal with advancing the theoretical properties of structure-preserving integrators. In this article, we derive a unified and modular integrator and develop methods for the numerically efficient implementation of such an integrator.}

More specifically, the contribution of this article to maximal-coordinate simulators is as follows. Firstly, we derive a variational integrator in maximal coordinates for physically accurate simulations, including good energy behavior and rigid, non-drifting constraints. While variational integrators are generally not new, we derive this integrator in a unified framework for typical dynamics components, including rigid bodies, joints, contacts, friction, actuators and external forces, springs, and dampers. Secondly, we provide an efficient graph-based solver algorithm for the system of nonlinear equations that form the integrator. The solver exploits the sparsity in the nonlinear system of equations to account for the increased number of variables in maximal coordinates. It achieves linear \myupdate{computational} complexity in the number of links and joints for mechanisms without kinematic loops and reduces the complexity for mechanisms with kinematic loops. For environment contacts and friction, the solver also achieves linear \myupdate{computational} complexity in the number of links and contact points while reducing the complexity for inter-mechanism contacts. Besides these theoretical contributions for efficient maximal-coordinate simulators based on variational integrators, we also provide an open-source implementation of such a simulator (see Section \ref{sec:evaluation}), which achieves competitive timing results compared to state-of-the-art simulators.

This article is structured as follows. In Section \ref{sec:simulator_components}, the desired simulator components are mathematically formalized. Based on these components, the variational integrator is derived in Section \ref{sec:math_integrator}, resulting in a system of nonlinear equations. The solver for this system of equations is presented in Section \ref{sec:numerical_solver}. An evaluation of the theoretical and numerical properties and application examples are given in Section \ref{sec:evaluation}, and conclusions are drawn in Section \ref{sec:conclusions}. The appendices provide background information, including our quaternion notation in Appendix \ref{sec:app_quaternions}. 

\section{Dynamics Components}\label{sec:simulator_components}
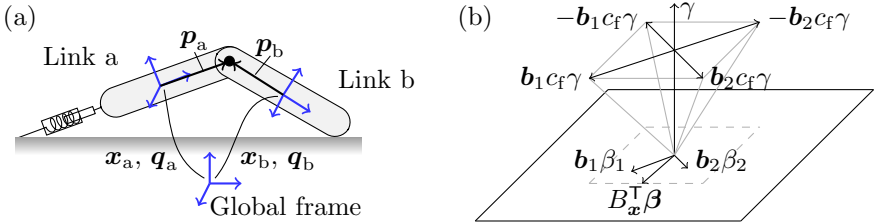
\begin{figure}[!hbt]
	\centering
	\begin{tikzpicture}
	\node[inner sep=0pt] (ref) at (0,0){};


	\coordinate (fc) at (5,-1.7);
	\draw[-,draw=black] ($(fc)+(0,0)$) -- ($(fc)+(1.75,1.75)$);
	\draw[-,draw=black] ($(fc)+(0,0)$) -- ($(fc)+(3.5,0)$);
	\draw[-,draw=black] ($(fc)+(3.5,0)$) -- ($(fc)+(5.25,1.75)$);
	\draw[-,draw=black] ($(fc)+(1.75,1.75)$) -- ($(fc)+(5.25,1.75)$);

	\draw[-,dashed,draw=black,draw=black!30] ($(fc)+(1.5,0.5)$) -- ($(fc)+(2.25,1.25)$);
	\draw[-,dashed,draw=black,draw=black!30] ($(fc)+(1.5,0.5)$) -- ($(fc)+(3.0,0.5)$);
	\draw[-,dashed,draw=black,draw=black!30] ($(fc)+(3.0,0.5)$) -- ($(fc)+(3.75,1.25)$);
	\draw[-,dashed,draw=black,draw=black!30] ($(fc)+(2.25,1.25)$) -- ($(fc)+(3.75,1.25)$);


	\draw[-,draw=black!30] ($(fc)+(1.5,1.9)$) -- ($(fc)+(2.25,2.65)$);
	\draw[-,draw=black!30] ($(fc)+(1.5,1.9)$) -- ($(fc)+(3.0,1.9)$);
	\draw[-,draw=black!30] ($(fc)+(3.0,1.9)$) -- ($(fc)+(3.75,2.65)$);
	\draw[-,draw=black!30] ($(fc)+(2.25,2.65)$) -- ($(fc)+(3.75,2.65)$);


	\draw[->,draw=black] ($(fc)+(2.625,0.875)$) -- ($(fc)+(2.625,2.9)$);

	\draw[-,draw=black!30] ($(fc)+(1.5,1.9)$) -- ($(fc)+(2.625,0.875)$);
	\draw[-,draw=black!30] ($(fc)+(3.0,1.9)$) -- ($(fc)+(2.625,0.875)$);
	\draw[-,draw=black!30] ($(fc)+(3.75,2.65)$) -- ($(fc)+(2.625,0.875)$);
	\draw[-,draw=black!30] ($(fc)+(2.25,2.65)$) -- ($(fc)+(2.625,0.875)$);

	\draw[->] ($(fc)+(2.625,2.275)$) -- ($(fc)+(1.5,1.9)$);
	\draw[->] ($(fc)+(2.625,2.275)$) -- ($(fc)+(3.75,2.65)$);
	\draw[->] ($(fc)+(2.625,2.275)$) -- ($(fc)+(2.25,2.65)$);
	\draw[->] ($(fc)+(2.625,2.275)$) -- ($(fc)+(3.0,1.9)$);

	\draw[->,draw=black] ($(fc)+(2.625,0.875)$) -- ($(fc)+(2.055,0.66)$);
	\draw[->,draw=black] ($(fc)+(2.625,0.875)$) -- ($(fc)+(2.8,0.7)$);
	\draw[->,draw=black] ($(fc)+(2.625,0.875)$) -- ($(fc)+(2.2,0.50)$);

	\node at ($(fc)+(1.0,1.9)$) {$\bno{b}{1}c_{\mathrm{f}}\gamma$};
	\node at ($(fc)+(3.5,1.9)$) {$\bno{b}{2}c_{\mathrm{f}}\gamma$};
	\node at ($(fc)+(1.6,2.7)$) {$-\bno{b}{1}c_{\mathrm{f}}\gamma$};
	\node at ($(fc)+(4.4,2.7)$) {$-\bno{b}{2}c_{\mathrm{f}}\gamma$};
	\node at ($(fc)+(1.65,0.8)$) {$\bno{b}{1}\beta_1$};
	\node at ($(fc)+(3.25,0.8)$) {$\bno{b}{2}\beta_2$};
	\node at ($(fc)+(2.1,0.275)$) {$B_{\bs{x}}\hT\bs{\beta}$};
	\node at ($(fc)+(2.8,2.8)$) {$\gamma$};

	\node[inner sep=0pt] (titleb) at (5.0,1.0){(b)};

	\node[inner sep=0pt] (titlea) at (-1.0,1.0){(a)};

	\draw[fill=black!5,rounded corners=6pt,rotate=20]
	(0,0.0) rectangle ++(2,-0.4);
	\draw[fill=black!5,rounded corners=6pt,rotate=-30]
	(1.12,1.45) rectangle ++(2,-0.4);
	
	\draw[fill=black] (1.76,0.43) circle (0.065);
	
	\coordinate (l1) at (0.85,0.1);
	\draw[->,draw=blue!80,thick] (l1) -- ($(l1)+(-0.16,0.4)$);
	\draw[->,draw=blue!80,thick] (l1) -- ($(l1)+(0.4,0.16)$);
	\draw[->,draw=blue!80,thick] (l1) -- ($(l1)+(-0.15,-0.3)$);

	\coordinate (l2) at (2.47,-0.02);
	\draw[->,draw=blue!80,thick] (l2) -- ($(l2)+(0.23,0.36)$);
	\draw[->,draw=blue!80,thick] (l2) -- ($(l2)+(0.36,-0.23)$);
	\draw[->,draw=blue!80,thick] (l2) -- ($(l2)+(-0.15,-0.3)$);

	\draw (-0.65,-0.45) -- (-1.05,-0.6);
	\draw (-0.032,-0.227) -- (0.07,-0.185);
	\draw[rotate=20] (-0.5,-0.15) rectangle ++(0.4,-0.1);
	\draw[rotate=20] (-0.8,-0.1) rectangle ++(0.5,-0.2);
	\draw[decoration={aspect=0.5, segment length=0.85mm, amplitude=0.90mm,coil},decorate] (-0.215,-0.29) -- (-0.68,-0.46); 
	
	\node [shading = axis,rectangle, left color=darkgray!50!white, right color=white, shading angle=0, minimum width = 130] at ($(l2)+(-1.25,-0.68)$) {};
	
	\draw[->,draw=black,thick] (l1) -- (1.72,0.41);
	\draw[->,draw=black,thick] (l2) -- (1.8,0.41);
	
	\coordinate (w) at (1.5,-1.2);
	\draw[->,draw=blue!80,thick] (w) -- ($(w)+(0.0,0.42)$);
	\draw[->,draw=blue!80,thick] (w) -- ($(w)+(0.42,0.0)$);
	\draw[->,draw=blue!80,thick] (w) -- ($(w)+(-0.15,-0.3)$);
	
	\node at (1.3,0.7) {$\btx{p}{a}$};
	\node at (2.3,0.6) {$\btx{p}{b}$};
	
	\node at (-0.2,0.5) {Link a};
	\node at (3.7,0.2) {Link b};
	
	\draw (1.3,0.55) -- (1.42,0.29);
	\draw (2.3,0.5) -- (2.1,0.25);
	
	\node at ($(w)+(1.0,-0.3)$) {Global frame};
	
	\draw ($(w)+(-0.07,0.07)$) to[out=150,in=-80 ] ($(l1)+(0.05,-0.05)$);
	\draw ($(w)+(0.07,0.07)$) to[out=50,in=180 ] ($(l2)+(-0.07,-0.02)$);
	
	\node at (0.6,-0.9) {$\xtx{a}$, $\qtx{a}$};
	\node at (2.4,-0.9) {$\xtx{b}$, $\qtx{b}$};
	\end{tikzpicture}
	\caption{Exemplary depiction of the simulator components. (a) A two-link mechanism with joints, a spring-damper, and contact subject to friction. (b) A four-sided linearized friction cone.}
	\label{fig:simulator_components}
\end{figure}

This section formulates the dynamics components for which the variational integrator is derived in Section \ref{sec:math_integrator}. Unit quaternions are used for rotations and orientations \myupdate{due to their computational efficiency} (see Appendix \ref{sec:app_quaternions} for our quaternion notation). \myupdate{However, other representations, such as rotation matrices, of which the group of unit quaternions is a double cover \cite{sola_micro_2018}}, could be used as well. Figure \ref{fig:simulator_components} shows a mechanism with the treated components.

\paragraph*{Rigid Body} 
In maximal coordinates, each of the $n_\mathrm{b}$ rigid bodies in a mechanism has a position $\bs{x}\in\mathbb{R}^{3}$ and orientation $\bs{q}\in\myupdate{\mathbb{H}}$ ($\bs{q}^{\mathrm{T}}\bs{q}=1$), as well as a translational velocity $\bs{v}\in\mathbb{R}^{3}$ and angular velocity $\bs{\omega}\in\mathbb{R}^{3}$. The configuration of a body is denoted as $\bs{z} = [\bs{x}\hT ~ \bs{q}\hT]\hT$, and the velocity is denoted as $\dot{\bs{z}} = [\bs{v}\hT ~ \bs{\omega}\hT]\hT$. Each body has a mass $m\in\mathbb{R}$ and a symmetric moment of inertia matrix $\bs{J}\in\mathbb{R}^{3\times3}$. All quantities refer to the center of mass of a body. Given $\bs{M}=m\bs{I}_{3\times 3}$, the kinetic energy of each body is $\mathcal{T}(\dot{\bs{z}}) = \frac{1}{2}\bs{v}\hT \bs{M}\bs{v} + \frac{1}{2}\bs{\omega}\hT \bs{J}\bs{\omega}$.

\paragraph*{Conservative Forces, Potentials, and Springs} 
Conservative forces in the dynamics are derived from potential functions $\mathcal{V}(\btx{z}{a}, \btx{z}{b}, \btx{z}{c}, \cdots)\in\mathbb{R}$ involving one or multiple bodies. Such potentials can represent, for example, gravity or springs.

\paragraph*{Non-conservative Forces, Actuators, and Dampers} 
Non-conservative forces, including actuators and dampers, are directly added to the dynamics as external forces $\bs{\mathrm{f}}\in\mathbb{R}^{3}$ or torques $\bs{\tau}\in\mathbb{R}^3$. Forces are described in the global frame and torques in the body frame. The wrench on a body is denoted as $\bs{\mathrm{w}} = [\bs{\mathrm{f}}\hT ~ \bs{\tau}\hT]\hT$. These wrenches are added for each body individually. Actuators fixed at joints are formulated by expressing their wrench in the frames of the connected bodies. 

As an example, for an actuator with wrench $\bs{\mathrm{w}}_{\mathrm{act}} = [\bs{\mathrm{f}}_{\mathrm{act}}\hT ~ \bs{\tau}_{\mathrm{act}}\hT]\hT$ at a joint between bodies a (parent) and b (child), the resulting wrenches for the bodies are
\begin{subequations}
	\begin{align}
		\bs{\mathrm{w}}_{\mathrm{a}} &= \begin{bmatrix}
			\mrmpreu{N}{\btx{\mathrm{f}}{a}} \\
			\mrmpreu{A}{\btx{\tau}{a}}
		\end{bmatrix} = 
		-\begin{bmatrix}
			\mrmpreu{N}{\btx{\mathrm{f}}{act}} \\
			\mrmpreu{A}{\btx{\tau}{act}} + \mrmpreu{A}{\btx{p}{a}}\times\mrmpreu{A}{\btx{\mathrm{f}}{act}}
		\end{bmatrix}, \\
		\bs{\mathrm{w}}_{\mathrm{b}} &= \begin{bmatrix}
			\mrmpreu{N}{\btx{\mathrm{f}}{b}} \\
			\mrmpreu{B}{\btx{\tau}{b}}
		\end{bmatrix} = 
		\begin{bmatrix}
			\mrmpreu{N}{\btx{\mathrm{f}}{act}} \\
			\mrmpreu{B}{\btx{\tau}{act}} + \mrmpreu{B}{\btx{p}{b}}\times\mrmpreu{B}{\btx{\mathrm{f}}{act}}
		\end{bmatrix},
	\end{align}
\end{subequations}
where $\mathrm{N}$ is the global frame, $\mathrm{A}$ and $\mathrm{B}$ are reference frames of bodies a and b, respectively, and $\bs{p}$ is a vector from center of mass to actuator.

\paragraph*{Joints and Equality Constraints} 
The joints of a mechanism consisting of one or multiple rigid bodies are represented by differentiable equality constraints $\bs{g}(\btx{z}{a}, \btx{z}{b}, \btx{z}{c}, \cdots)=\bs{0}\in\mathbb{R}^{n_\mathrm{e}}$, where $n_\mathrm{e}$ is the number of equality constraints on the mechanism. Typically, equality constraints are formulated for all $i$ kinematically connected pairs of rigid bodies independently and subsequently stacked into one constraint function $\bs{g} = [\bs{g}_1\hT \cdots \bs{g}_i\hT]\hT$. In maximal coordinates, two generic equality constraint functions, one for the translational and one for the rotational movement of the two connected bodies, can be combined to create most of the common joints encountered in mechanisms. This insight greatly simplifies analytic gradient calculations for fast computations and gives direct access to minimal coordinates as well (see Appendix \ref{sec:app_joints} for a list of possible joints \myupdate{and the recovery of minimal coordinates}).

\paragraph*{Contacts and Inequality Constraints}
Rigid contacts between multiple bodies and with the environment are represented by differentiable inequality constraints $\bs{\phi}(\btx{z}{a}, \btx{z}{b}, \btx{z}{c}, \cdots)\geq\bs{0}\in\mathbb{R}^{n_{\mathrm{i}}}$, where $n_\mathrm{i}$ is the number of inequality constraints on the mechanism. As for joints, inequality constraints are typically formulated independently for all $j$ pairs of rigid bodies from their signed distance function and subsequently stacked into one constraint function $\bs{\phi} = [\phi_1\hT \cdots \phi_j\hT]\hT$. 

As an example, ground contact of a single point contact on a body in maximal coordinates always has the form $\phi(\bs{z}) = \bs{e}_{\mathrm{z}}\hT(\bs{x} + \bs{q}\cdot\bs{p}\cdot\bs{q}^{-1}) \geq 0$, where $\bs{e}_{\mathrm{z}}$ is the z-axis unit vector in the global frame and $\bs{p}\in\mathbb{R}^{3}$ points from center of mass to contact point in the body's frame. As before, this consistent structure allows for the calculation of analytic gradients to improve computation time.

\myupdate{\paragraph*{Static and Sliding Friction}
In the dynamics, static and sliding friction are added as external forces on the bodies. These forces are calculated by solving an optimization problem derived from the maximum dissipation principle \cite{preclik_maximum_2018}. We are using a linearized friction cone \cite{stewart_implicit_1996}, although a nonlinear friction cone could be used as well. The maximum dissipation principle states that the energy dissipation rate of the bodies in contact is maximized by a friction force $\bs{\beta}\in\mathbb{R}^{n_\mathrm{c}n_\mathrm{f}}$, where $n_\mathrm{c}$ is the number of contacts and $n_\mathrm{f}$ is the even number of basis vectors of the friction cone. The basis vectors $\bno{b}{i}\in\mathbb{R}^{3}$ of the linearized friction cone are depicted in Fig. \ref{fig:simulator_components} (b). A detailed derivation of the optimization problem is stated in Appendix \ref{sec:app_friction}. The optimization problem resulting from the maximum dissipation principle for the bodies of a mechanism is
\begin{subequations}\label{eqn:max_dissip}
	\begin{alignat}{2}
		&\min_{\bs{\beta}} ~ &&\begin{bmatrix}\dot{\bs{z}}_{\mathrm{a}}\hT & \dot{\bs{z}}_{\mathrm{b}}\hT & \dot{\bs{z}}_{\mathrm{c}}\hT & \cdots\end{bmatrix} \bs{B}\left(\btx{z}{a},\btx{z}{b},\btx{z}{c},\cdots\right)\hT \bs{\beta},\\
		&\text{s.t.}~&&\bs{E}\hT\bs{\beta} \leq \bs{C}_\mathrm{f}\bs{\gamma},\\
		& &&\bs{\beta}\geq \bs{0},
	\end{alignat}
\end{subequations}
where $\bs{B}\in\mathbb{R}^{6n_\mathrm{b}\times n_\mathrm{c}n_\mathrm{f}}$ maps the $n_\mathrm{f}$-dimensional friction forces at each of the $n_\mathrm{c}$ contact point to a six-dimensional wrench on the respective bodies. The constraint $\bs{E}\hT\bs{\beta} \leq \bs{C}_\mathrm{f}\bs{\gamma}$ describes the limit on the friction forces with $\bs{E}=\mathrm{diag}(\pmb{1}, \cdots, \pmb{1})\in\mathbb{R}^{n_\mathrm{c}n_\mathrm{f}\times n_\mathrm{c}}$, $\pmb{1} = [1 ~ \cdots ~ 1]\hT\in\mathbb{R}^{n_\mathrm{f}}$, normal forces $\bs{\gamma}\in\mathbb{R}^{n_\mathrm{c}}$, and the friction coefficient matrix $\bs{C}_\mathrm{f} = \mathrm{diag}(c_{\mathrm{f},1},\cdots,c_{\mathrm{f},n_\mathrm{c}})\in\mathbb{R}^{n_\mathrm{c}\times n_\mathrm{c}}$ containing the friction coefficients $c_\mathrm{f}$ for each contact point. The friction wrenches for each body resulting from the optimization are added as external wrenches to the dynamics.}

\myupdate{As an example, the optimization problem for ground contact of a single point contact on a body in maximal coordinates always has the form
\begin{subequations}\label{eqn:max_dissip_expl}
	\begin{alignat}{2}
		&\min_{\bs{\beta}} ~ &&\dot{\bs{z}}\hT \bs{B}\hT \bs{\beta},\\
		&\text{s.t.}~&&\pmb{1}\hT\bs{\beta} \leq c_\mathrm{f}\gamma,\\
		& &&\bs{\beta}\geq \bs{0},
	\end{alignat}
\end{subequations}
with $\bs{B}\hT=[\bs{B}_{\bs{x}} ~ \bs{B}_{\bs{q}}]\hT$ consisting of 
\begin{subequations}
	\begin{align}
		\bs{B}_{\bs{x}}\hT &= \begin{bmatrix}
			\bs{b}_1 ~ -\bs{b}_1 ~ \cdots ~ \bs{b}_{\frac{n_\mathrm{f}}{2}} ~ -\bs{b}_{\frac{n_\mathrm{f}}{2}}
		\end{bmatrix}\in\mathbb{R}^{3\times n_\mathrm{f}},\\
		\bs{B}_{\bs{q}}\hT &= \bs{p}^\times \bs{Q}(\bs{q})\bs{B}_{\bs{x}}\hT\in\mathbb{R}^{3\times n_\mathrm{f}},
	\end{align}
\end{subequations}
where the vector from the body's center of mass to the contact point is denoted $\bs{p}$, the $^{\times}$ operator creates the skew-symmetric matrix from this vector, and $\bs{Q}(\bs{q})$ is the rotation matrix for quaternion $\bs{q}$.}

\section{Mathematical Integrator}\label{sec:math_integrator}
A first-order variational (symplectic) integrator \cite{marsden_discrete_2001} is derived for the simulator components described in the previous section. This integrator discretizes the rigid-body dynamics while maintaining energy and momentum conservation properties as well as constraint satisfaction. Higher-order variational integrators are possible \cite{wenger_constrained_2017,wenger_construction_2017}, and we restrict the derivation to first order for clarity. First, the derivation for unconstrained dynamics is provided. Afterward, equality and inequality constraints are added to the integrator, and finally, friction dynamics are incorporated.

\subsection{Unconstrained Dynamics}
The derivation of the integrator for unconstrained dynamics is split into translational and rotational components for clarity. Note that the derivation also holds for coupled translational and rotational dynamics. Variational integrators are based on the principle of least action, which states that a mechanical system takes the path of least action when going from a fixed start point to a fixed end point. \myupdate{External forces and torques are incorporated with the Lagrange-d'Alembert principle.} Action has the dimensions $[\text{Energy}]\times[\text{Time}]$ and the unconstrained action integral $S_0$ \myupdate{with } is defined as 
\begin{equation}\label{eqn:unconstrained_action}
	S_0(\bs{z},\dot{\bs{z}}) = \int_{t_0}^{t_N} \mathcal{L}(\bs{z},\dot{\bs{z}}) ~ \mathrm{d}t + \int_{t_0}^{t_N} \begin{bmatrix}\bs{\mathrm{f}}\hT & 2\Lmat{\bs{q}}\bs{V}\hT\bs{\tau}\end{bmatrix}\hT \bs{z} ~ \mathrm{d}t,
\end{equation}
where $\mathcal{L} = \mathcal{T}-\mathcal{V}$ is the Lagrangian with kinetic energy $\mathcal{T}$ and potential energy $\mathcal{V}$. A brief explanation of the \myupdate{external force and torque} component in \eqref{eqn:unconstrained_action}, as well as our quaternion notation, are given in Appendix \ref{sec:app_quaternions}. Further details on virtual work for quaternions can be found in \cite{baruh_analytical_1999,shivarama_hamilton_2004}.

\paragraph*{Translational Component}
The translational component of the action integral \eqref{eqn:unconstrained_action} is
\begin{equation}\label{eqn:unconstrained_translational}
	S_{0,\mathrm{T}}(\bs{x},\bs{v}) = \int_{t_0}^{t_N}\mathcal{L}_\mathrm{T}(\bs{x},\bs{v}) ~ \mathrm{d}t + \int_{t_0}^{t_N}\bs{\mathrm{f}}\hT \bs{x} ~ \mathrm{d}t.
\end{equation}

For numerical integration, \eqref{eqn:unconstrained_translational} is discretized.  A first-order discretization of the integral and a first-order approximation of the velocity, 
\begin{equation}\label{eqn:vk}
\vno{k} = \frac{\xno{k+1}-\xno{k}}{\dt},
\end{equation}
with step size $\Delta t$ is used to obtain the discrete action sum
\begin{equation}
	S_{\mathrm{d}, 0, \mathrm{T}}(\xno{k},\vno{k}) = \sum_{k=0}^{N-1}\left( \mathcal{L}_\mathrm{T}(\xno{k},\vno{k}) + \bs{\mathrm{f}}_{k}\hT \xno{k}\right) \dt.
\end{equation}

For a first-order integrator, the principle of least action must be fulfilled for trajectories consisting of three knot points, i.e., from $0$ to $N=2$. Since $\xno{0}$ and $\xno{2}$ are fixed start and end points of the trajectory, only $\xno{1}$ can vary. Therefore, the action sum is minimized with respect to the position $\xno{1}$:
\begin{equation}\label{eqn:unconstrained_derivative}
	\nabla_{\xno{1}}S_{\mathrm{d}, 0, \mathrm{T}} = -\btx{d}{0,T}\Delta t = \bs{0},
\end{equation}
where $\btx{d}{0,T}$ are the resulting implicit discretized translational dynamics. In words, if equation \eqref{eqn:unconstrained_derivative} is fulfilled, we have found the discrete approximation of the physically correct trajectory consisting of the knot points $\xno{0}$, $\xno{1}$, and $\xno{2}$. Note that the derivative with respect to $\bno{x}{1}$ is taken for each body in a mechanism.

An example for a single body with potential function $\mathcal{V}(\bs{x})$ yields
\begin{align}\label{eqn:unconstrained_example}
	\btx{d}{0,T}(\vno{1}) &= -\nabla_{\xno{1}}\left(\mathcal{L}_\mathrm{T}(\xno{0},\vno{0}) + \bs{\mathrm{f}}_{0}\hT \xno{0} + \mathcal{L}_\mathrm{T}(\xno{1},\vno{1}) + \bs{\mathrm{f}}_{1}\hT \xno{1}\right)\nonumber\\
	&= -\left(\bs{M}\frac{\xno{1}-\xno{0}}{\dt} - \bs{M}\frac{\xno{2}-\xno{1}}{\dt} - \nabla_{\xno{1}}\mathcal{V}(\xno{1}) + \bs{\mathrm{f}}_{1}\right)\nonumber\\
	&= \bs{M}\frac{\vno{1}-\vno{0}}{\dt} + \nabla_{\xno{1}}\mathcal{V}(\xno{1}) -\bs{\mathrm{f}}_{1} = \bs{0},
\end{align}
which resembles the discretized version of Newton's second law, $\bs{M}\dot{\bs{v}}-\bs{\mathrm{f}}=\bs{0}$.

The physically accurate dynamics are obtained by varying $\xno{1}$ with fixed $\xno{0}$ and $\xno{2}$. However, when integrating dynamics forward in time, we start from a known initial state $\xno{0}$ and $\vno{0}$.
From \eqref{eqn:vk}, the position at the next time step, $\xno{1}$, is calculated as
\begin{equation}\label{eqn:updatePos}
	\xno{1} = \xno{0} + \vno{0}\dt.
\end{equation}
Then, the implicit dynamics equations \eqref{eqn:unconstrained_derivative} are solved to obtain the velocity $\vno{1}$. \myupdate{This resulting integration scheme consisting of \eqref{eqn:updatePos} and \eqref{eqn:unconstrained_derivative} is the symplectic Euler method \cite{hairer_geometric_2006}}.

\paragraph*{Rotational Component}
The rotational component of the integrator can be derived similarly to the translation case. A description for an unconstrained floating single rigid body is presented in \cite{manchester_quaternion_2016}, and we later extend the derivation to constrained multibody systems.

The action integral for the rotational component is
\begin{equation}\label{eqn:action_rot}
	S_{0,\mathrm{R}}(\bs{q},\bs{\omega}) = \int_{t_0}^{t_N} \mathcal{L}_\mathrm{R}(\bs{q},\bs{\omega})~\mathrm{d}t + \int_{t_0}^{t_N}2\bs{\tau}\hT \bs{V} \LTmat{\bs{q}} \bs{q}~\mathrm{d}t.
\end{equation}

To maintain unit norm in the quaternion update (see Appendix \ref{sec:app_quaternions}), the discrete quaternion angular velocity is defined as
\begin{equation}
	\bar{\bs{\omega}}_k = \begin{bmatrix}\sqrt{\left(\frac{2}{\dt}\right)^2 - \bno{\omega}{k}\hT\bno{\omega}{k}}~\\\bno{\omega}{k}\end{bmatrix} = \myupdate{\frac{2}{\dt}\Lmat{\bno{q}{k}}\hT\bno{q}{k+1}}.
\end{equation}

As before, the action integral \eqref{eqn:action_rot} is discretized to obtain the action sum
\begin{equation}
	S_{\mathrm{d},0,\mathrm{R}}(\qno{k},\wno{k}) = \sum_{k=0}^{N-1}\left(\mathcal{L}_\mathrm{R}(\qno{k},\wno{k}) + 2\bs{\tau}_k\hT \bs{V} \LTmat{\qno{k}} \qno{k}\right) \dt.
\end{equation}

The principle of least action is fulfilled by minimizing the discrete action sum from $0$ to $N=2$ over the orientation $\qno{1}$:
\begin{equation}\label{eqn:derivDiscRot}
\nabla^{\mathrm{r}}_{\qno{1}}S_{\mathrm{d},0,\mathrm{R}} = -\btx{d}{0,R}\Delta t = \bs{0},
\end{equation}
where $\btx{d}{0,R}$ are the implicit discretized rotational dynamics. Note that we have used the rotational gradient $\nabla^{\mathrm{r}}$ (see Appendix \ref{sec:app_quaternions}) and that the derivative with respect to $\qno{1}$ is taken for each body in a mechanism.

An example for a single body with potential function $\mathcal{V}(\bs{q})$ yields
\begin{multline}\label{eqn:dR1}
	\btx{d}{0,R}(\wno{1}) =  \bs{J} \wno{1}\sqrt{\tfrac{4}{\dt^2}-\wno{1}\hT\wno{1}} + \wno{1}^{\times} \bs{J} \wno{1} - \\ \bs{J} \wno{0}\sqrt{\tfrac{4}{\dt^2}-\wno{0}\hT\wno{0}} + \wno{0}^{\times} \bs{J} \wno{0} + \nabla^{\mathrm{r}}_{\qno{1}}\mathcal{V}(\qno{1}) -2\bno{\tau}{2} = \bs{0},
\end{multline}
which---analogous to the translational case---bears resemblance to Euler's equations for rotations $J\dot{\bs{\omega}}+\bs{\omega}^{\times}J\bs{\omega} - \bs{\tau} = \bs{0}$.

An integration step given $\bs{q}_0$ and $\bs{\omega}_0$ is performed by first calculating
\begin{equation} \label{eqn:updateOr}
	\qno{1} = \frac{\dt}{2}\Lmat{\qno{0}}\bar{\bs{\omega}}_0,
\end{equation}
and subsequently solving \eqref{eqn:derivDiscRot} for the angular velocity $\bs{\omega}_1$.

\subsection{Equality Constrained Dynamics}
Equality constraint functions $\bs{g}(\bs{z})$ with \myupdate{Lagrange multiplier} $\bs{\lambda}\in\mathbb{R}^{n_\mathrm{e}}$ are added to the integrator by appending them to the action integral:
\begin{equation}
	S(\bs{z},\dot{\bs{z}},\bs{\lambda}) = S_0(\bs{z},\dot{\bs{z}}) + \int_{t_0}^{t_N}\bs{\lambda}\hT \bs{g}(\bs{z}) ~ \mathrm{d}t.
\end{equation}

Accordingly, the discrete action sum changes to
\begin{equation}\label{eqn:constrained_discrete}
	S_{\mathrm{d}}(\bno{z}{k},\dot{\bs{z}}_k,\bno{\lambda}{k}) = S_{\mathrm{d},0}(\bno{z}{k},\dot{\bs{z}}_k)+\sum_{k=0}^{N-1}\lano{k}\hT \bs{g}(\bno{z}{k}) \dt.
\end{equation}

Taking the gradient of \eqref{eqn:constrained_discrete} with respect to $\bno{x}{1}$ and $\bno{q}{1}$ yields the constrained implicit discretized dynamics
\begin{subequations}\label{eqn:constrained_d}
	\begin{align}
		\bs{d}(\dot{\bs{z}}_1,\bno{\lambda}{1}) = \bno{d}{0}(\dot{\bs{z}}_1) - \bs{G}(\bno{z}{1})\hT\bno{\lambda}{1} &= \bs{0}, \\
		\bs{g}\left(\bno{z}{2}\left(\dot{\bs{z}}_1\right)\right) &= \bs{0},
	\end{align}
\end{subequations}
where
\begin{equation}
	\bs{G}(\bs{z}) = \begin{bmatrix}
		\frac{\partial \bs{g}(\bs{z})}{\partial \bs{x}} & \frac{\partial \bs{g}(\bs{z})}{\partial^\mathrm{r} \bs{q}}
	\end{bmatrix}.
\end{equation}

\myupdate{Physically, the constraint forces $\myupdate{\bs{G}(\bno{z}{1})\hT}\bno{\lambda}{1}$ act on the rigid bodies to guarantee satisfaction of constraints $\bs{g}=\bs{0}$. Mathematically, $\bs{\lambda}$ serves a similar purpose as Lagrange multipliers in constrained optimization.}

The integration step starting from $\bs{z}_0$ and $\dot{\bs{z}}_0$ is calculated as follows. First, $\bs{z}_1$ is calculated from the update rules \eqref{eqn:updatePos} and \eqref{eqn:updateOr}. Then, the nonlinear system of equations \eqref{eqn:constrained_d} is solved for $\dot{\bs{z}}_1$ and $\bno{\lambda}{1}$. Note that the constraints are fulfilled for $\bno{z}{2}$ which depends on $\dot{\bs{z}}_1$ through update rules \eqref{eqn:updatePos} and \eqref{eqn:updateOr}. Therefore, the resulting velocity $\dot{\bs{z}}_1$ always ensures constraint satisfaction for the next position $\bs{z}_2$.

\subsection{Inequality Constrained Dynamics}
Inequality constraint functions $\bs{\phi}(\bs{z})$ with \myupdate{Lagrange multipliers} $\bs{\gamma}\in\mathbb{R}^{n_\mathrm{i}}$ are added to the integrator in a similar fashion. Physically, the \myupdate{multipliers} $\bs{\gamma}$ are the \myupdate{magnitudes of the normal} forces at the contacts. To add the constraints to the dynamics, they are discretized and formulated as a nonlinear complementarity problem (NCP)
\begin{subequations}\label{eqn:contact_comp}
	\begin{align}
		\bs{\phi}(\bs{z}_k) &\geq \bs{0},\\
		\bno{\gamma}{k} &\geq \bs{0},\\
		\bs{\phi}(\bno{z}{k})\hT\bno{\gamma}{k} &= \bs{0},
	\end{align}
\end{subequations}
with element-wise $\geq$, for which we use the standard shorthand notation
\begin{equation}
	\bs{0}\leq\bs{\phi}(\bs{z}_k)\perp\bno{\gamma}{k}\geq\bs{0}.
\end{equation}

The resulting dynamics are
\begin{subequations}\label{eqn:ineqconstrained_d}
	\begin{align}
		\bs{d}(\dot{\bs{z}}_1,\bno{\gamma}{1}) = \bno{d}{0}(\dot{\bs{z}}_1) - \bs{N}(\bno{z}{1})\hT\bno{\gamma}{1} &= \bs{0}, \\
		\bs{0}\leq\bs{\phi}(\bs{z}_2(\dot{\bs{z}}_1))\perp\bno{\gamma}{1}&\geq\bs{0},
	\end{align}
\end{subequations}
where
\begin{equation}
	\bs{N}(\bs{z}) = \begin{bmatrix}
		\frac{\partial \bs{\phi}(\bs{z})}{\partial \bs{x}} & \frac{\partial \bs{\phi}(\bs{z})}{\partial^\mathrm{r} \bs{q}}
	\end{bmatrix}.
\end{equation}

The integration step starting from $\bs{z}_0$ and $\dot{\bs{z}}_0$ is again performed by first calculating $\bs{z}_1$ from \eqref{eqn:updatePos} and \eqref{eqn:updateOr}, and then solving \eqref{eqn:ineqconstrained_d} for $\dot{\bs{z}}_1$ and $\bno{\gamma}{1}$.

\subsection{Friction Dynamics}
The friction dynamics are included in the variational integrator by discretizing the maximum dissipation principle \eqref{eqn:max_dissip}:
\myupdate{\begin{subequations}\label{eqn:max_dissip_disc}
	\begin{alignat}{2}
		&\min_{\bs{\beta}} ~ &&\dot{\bs{z}}_{k}\hT \bs{B}(\bno{z}{k})\hT \bno{\beta}{k},\\
		&\text{s.t.}~&&\bs{E}\hT\bno{\beta}{k} \leq \bs{C}_\mathrm{f}\bno{\gamma}{k},\\
		& &&\bno{\beta}{k}\geq \bs{0}.
	\end{alignat}
\end{subequations}
Note that the normal force multipliers $\bs{\gamma}$ will result from contact inequality constraints in the form of \eqref{eqn:contact_comp}.}

\myupdate{As with the contact constraint, we formulate \eqref{eqn:max_dissip_disc} as an NCP with Lagrange multipliers $\bno{\psi}{k}\in\mathrm{R}^{n_\mathrm{c}}$---the tangential velocities at the contact points---and $\bs{\eta}_k\in\mathbb{R}^{n_\mathrm{c}n_\mathrm{f}}$:
\begin{subequations}\label{eqn:fric_comp}
	\begin{align}
		\bs{B}(\bs{z}_{k})\dot{\bs{z}}_{k} + \bs{E}\bno{\psi}{k} - \bs{\eta}_k &= \bs{0},\\
		\bs{0} \leq \bs{C}_\mathrm{f}\bno{\gamma}{k} - \bs{E}\hT\bs{\beta}_k \perp \bno{\psi}{k} &\geq \bs{0},\\
		\bs{0} \leq \bs{\beta}_k \perp \bs{\eta}_k &\geq \bs{0}.
	\end{align}
\end{subequations}}

\myupdate{Given a mechanism with $n_\mathrm{c}$ contact inequality constraints with friction, the resulting dynamics are
\begin{subequations}\label{eqn:friction_d}
	\begin{align}
		\bs{d}(\dot{\bs{z}}_1,\bno{\gamma}{1},\bno{\beta}{1},\bno{\psi}{1},\bno{\eta}{1}) = \bno{d}{0}(\dot{\bs{z}}_1) - \bs{N}(\bno{z}{1})\hT\bno{\gamma}{1} - \bs{B}(\bno{z}{1})\hT\bno{\beta}{1} &= \bs{0}, \\
		\bs{0}\leq\bs{\phi}(\bs{z}_2(\dot{\bs{z}}_1))\perp\bno{\gamma}{1}&\geq\bs{0}, \\
		\bs{B}(\bs{z}_{1})\dot{\bs{z}}_{1} + \bs{E}\bno{\psi}{1} - \bs{\eta}_1 &= \bs{0},\\
		\bs{0} \leq \bs{C}_\mathrm{f}\bno{\gamma}{1} - \bs{E}\hT\bs{\beta}_1 \perp \bno{\psi}{1} &\geq \bs{0},\\
		\bs{0} \leq \bs{\beta}_1 \perp \bs{\eta}_1 &\geq \bs{0},
	\end{align}
\end{subequations}
and the integration step starting from $\bs{z}_0$ and $\dot{\bs{z}}_0$ is performed by first calculating $\bs{z}_1$ from \eqref{eqn:updatePos} and \eqref{eqn:updateOr}, and then solving \eqref{eqn:friction_d} for $\dot{\bs{z}}_1$, $\bno{\gamma}{1}$, $\bno{\beta}{1}$, $\bno{\psi}{1}$, and $\bno{\eta}{1}$.}

\subsection{Complete Integrator}
Putting all components together, the simulation of one time step given $\bs{z}_0$ and $\dot{\bs{z}}_0$ is performed by first calculating $\bs{z}_1$ from \eqref{eqn:updatePos} and \eqref{eqn:updateOr}. Subsequently, the constrained implicit dynamics must be solved:
\begin{subequations}\label{eqn:complete_integrator}
	\begin{align}
		\bs{d}(\dot{\bs{z}}_1,\bno{\lambda}{1},\bno{\gamma}{1},\bno{\beta}{1}\myupdate{,\bno{\psi}{1},\bno{\eta}{1}}) = \bno{d}{0}(\dot{\bs{z}}_1) - \bs{G}\hT\bno{\lambda}{1} - \bs{N}\hT\bno{\gamma}{1} - \bs{B}\hT\bno{\beta}{1} &= \bs{0}, \\
		\bs{g}\left(\bs{z}_2(\dot{\bs{z}}_1)\right) &= \bs{0},\\
		\bs{0}\leq\bs{\phi}\left(\bs{z}_2(\dot{\bs{z}}_1)\right)\perp\bno{\gamma}{1}&\geq\bs{0},\\
		\bs{B}\dot{\bs{z}}_1 + \bs{E}\bno{\psi}{1} - \bs{\eta}_1 &= \bs{0},\label{eqn:fric1}\\
		\myupdate{\bs{0} \leq \bs{C}_\mathrm{f}\bno{\gamma}{1} - \bs{E}\hT\bs{\beta}_1 \perp \bno{\psi}{1}} &\myupdate{\geq \bs{0}},\label{eqn:fric2}\\
		\bs{0} \leq \bs{\beta}_1 \perp \bs{\eta}_1 &\geq \bs{0}\label{eqn:fric3}.
	\end{align}
\end{subequations}
Note that \eqref{eqn:complete_integrator} contains the equations and constraints for all bodies of a mechanism. 

\myupdate{If the initial velocity $\dot{\bs{z}}_0$ is unknown, and a non-constraint-fulfilling one is chosen, an error will be incurred for the first time step. The magnitude of the error depends on the magnitude of the initial constraint violation. The discrete Legendre transform can be used to determine a constraint-fulfilling initial velocity. We refer to the extensive explanation in \cite{leyendecker_variational_2008} for details on initializing a simulation with the Legendre transform.}

\section{Numerical Solver}\label{sec:numerical_solver}
The variational integrator \eqref{eqn:complete_integrator} can be summarized as a system of nonlinear equations with inequality constraints:
\begin{subequations}\label{eqn:system}
	\begin{align}
	\bs{f}(\bs{s})&=\bs{0}, \\
	\bs{h}(\bs{s})&\geq\bs{0},
	\end{align}
\end{subequations}
\myupdate{with solution vector} $\bs{s} = [\dot{\bs{z}}_1\hT ~ \bno{\lambda}{1}\hT ~ \bno{\gamma}{1}\hT ~ \bno{\beta}{1}\hT ~ \bs{\psi}_1\hT ~ \bs{\eta}_1\hT]\hT$. In this section, the algorithms derived for solving the system \eqref{eqn:system} are applicable to the class of Newton-based root-finding methods. Certain methods of this class, for example interior-point methods \cite{nocedal_numerical_2006}, introduce slack variables \myupdate{$\bs{\sigma}$ and additional constraints $\bs{h}(\bs{s})=\bs{\sigma}$ to facilitate the numerical treatment of the inequality constraints. Since the slack variables match the inequality constraints up to the desired solution tolerance, the theoretical properties of the integrator still hold. The solution vector $\bs{s}$ and the system $\bs{f}(\bs{s})$ would be extended by these slack variables and constraints, but the computational complexity does not change if each inequality constraint and its associated slack constraint are considered as a single node in a graph.} Moreover, the graph-based argument is not limited to mechanical systems and can be implemented for any graph-based system, but we will restrict the discussion to mechanical systems.

At the core, Newton-based methods iteratively produce solution approximations for \eqref{eqn:system} with the procedure
\begin{equation}\label{eqn:iteration}
	\bs{s}^{(i+1)} = \bs{s}^{(i)} - \bs{F}(\bs{s}^{(i)})^{-1}\bs{f}(\bs{s}^{(i)}),
\end{equation}
where 
\begin{equation}
	\bs{F}(\bs{s}) = \frac{\partial \bs{f}(\bs{s})}{\partial \bs{s}}.
\end{equation}

Numerically, \eqref{eqn:iteration} is formulated as a linear system of equations
\begin{equation}\label{eqn:linear_system}
	\bs{F}(\bs{s}^{(i)})\Delta\bs{s}^{(i)} = -\bs{f}(\bs{s}^{(i)}),
\end{equation}
where the result $\Delta\bs{s}^{(i)}$ is used to obtain $\bs{s}^{(i+1)} = \bs{s}^{(i)} + \Delta\bs{s}^{(i)}$.

Linear systems of the form \eqref{eqn:linear_system} are solved with decomposition and backsubstitution with overall cubic \myupdate{computational} complexity $\mathcal{O}(n^3)$. For general integrators, including the variational integrator derived in the previous section, \eqref{eqn:linear_system} is neither symmetric nor block symmetric. Therefore, the LDU decomposition \cite{kwak_linear_2004} for unsymmetric systems is chosen as the foundation of the algorithms. While \eqref{eqn:linear_system} is unsymmetric, its sparsity pattern, i.e., the zero and non-zero entries, is block-symmetric, and the following algorithms exploit this sparsity to improve on the complexity.  

\subsection{Linear-Complexity Algorithm}
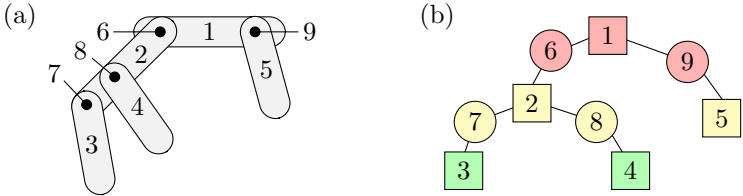
\begin{figure}[!htb]
	\centering
	\begin{tikzpicture}

    \coordinate (Ca) at (0.0,0.0);	
    \node at ($(Ca)+(-1.5,0.0)$) {(a)};
    
    \draw[fill=black!5,rounded corners=6pt]
    ($(Ca)+(0,0.0)$) rectangle ++(2,-0.4);
    \draw[fill=black!5,rounded corners=6pt,rotate=-45]
    ($(Ca)+(0.2,0.325)$) rectangle ++(0.4,-1.8);
    \draw[fill=black!5,rounded corners=6pt,rotate=10]
    ($(Ca)+(-1.015,-0.86)$) rectangle ++(0.4,-1.4);
    \draw[fill=black!5,rounded corners=6pt,rotate=35]
    ($(Ca)+(-0.87,-0.32)$) rectangle ++(0.4,-1.4);
    \draw[fill=black!5,rounded corners=6pt,rotate=15]
    ($(Ca)+(1.3,-0.4)$) rectangle ++(0.4,-1.4);
    
    \draw[fill=black] ($(Ca)+(0.35,-0.2)$) circle (0.065);
    \draw[fill=black] ($(Ca)+(-0.62,-1.17)$) circle (0.065);
    \draw[fill=black] ($(Ca)+(-0.25,-0.8)$) circle (0.065);
    \draw[fill=black] ($(Ca)+(1.6,-0.2)$) circle (0.065);
    
    \node at ($(Ca)+(1,-0.2)$) {1};
    \node at ($(Ca)+(0.1,-0.5)$) {2};
    \node at ($(Ca)+(-0.55,-1.7)$) {3};
    \node at ($(Ca)+(0.05,-1.2)$) {4};
    \node at ($(Ca)+(1.75,-0.7)$) {5};
    
    \draw ($(Ca)+(0.35,-0.2)$) -- ($(Ca)+(-0.27,-0.2)$);
    \draw ($(Ca)+(-0.62,-1.17)$) -- ($(Ca)+(-0.92,-0.85)$);
    \draw ($(Ca)+(-0.25,-0.8)$) -- ($(Ca)+(-0.55,-0.55)$);
    \draw ($(Ca)+(1.6,-0.2)$) -- ($(Ca)+(2.17,-0.2)$);
    
    \node at ($(Ca)+(-0.4,-0.2)$) {6};
    \node at ($(Ca)+(-1.05,-0.75)$) {7};
    \node at ($(Ca)+(-0.7,-0.45)$) {8};
    \node at ($(Ca)+(2.32,-0.2)$) {9};

    \node at (4,0) {(b)};
    
    \draw (6.2,-0.2) -- (5.5,-0.5);
    \draw (5.5,-0.45) -- (5.1,-1.2);
    \draw (5.3,-1.1) -- (4.5,-1.4);
    \draw (4.5,-1.4) -- (4.3,-2);
    \draw (5.2,-1.1) -- (6.1,-1.4);
    \draw (6.2,-1.4) -- (6.5,-1.9);
    \draw (6.2,-0.2) -- (7.3,-0.6);
    \draw (7.4,-0.6) -- (7.9,-1.3);

    \draw[fill=red!30] (6,0.0) rectangle ++(0.5,-0.5);
    \draw[fill=yellow!30] (5,-0.9) rectangle ++(0.5,-0.5);
    \draw[fill=green!30] (4.1,-1.8) rectangle ++(0.5,-0.5);
    \draw[fill=green!30] (6.3,-1.8) rectangle ++(0.5,-0.5);
    \draw[fill=yellow!30] (7.5,-1.1) rectangle ++(0.5,-0.5);
    
    \draw[fill=red!30] (5.5,-0.45) circle (0.275);
    \draw[fill=yellow!30] (4.5,-1.4) circle (0.275);
    \draw[fill=yellow!30] (6.1,-1.4) circle (0.275);
    \draw[fill=red!30] (7.3,-0.6) circle (0.275);
    
    \node at (6.25,-0.25) {1};
    \node at (5.25,-1.15) {2};
    \node at (4.35,-2.05) {3};
    \node at (6.55,-2.05) {4};
    \node at (7.75,-1.35) {5};
    \node at (5.5,-0.45) {6};
    \node at (4.5,-1.4) {7};
    \node at (6.1,-1.4) {8};
    \node at (7.3,-0.6) {9};
    \end{tikzpicture}
	\caption{(a) A mechanism with five links and four joints. (b) A graph representing the mechanism and its matrix. Squares represent links, circles represent joints. Coloring represents different levels of the tree-shaped graph.}
	\label{fig:mechanism}
\end{figure}

For mechanisms without kinematic loops, the LDU decomposition can be modified to obtain decomposition and backsubstitution with linear \myupdate{computational} complexity $\mathcal{O}(n)$, where $n$ is the number of nodes in the corresponding graph. This modification is achieved by taking into account the graph representing the components of a mechanism and its associated $\bs{F}$ matrix. Consider, for example, the mechanism and graph in Fig. \ref{fig:mechanism}.

According to \cite{duff_direct_2017}, the $\bs{F}$ matrix corresponding to an acyclic graph, i.e., a mechanism without kinematic loops, can be decomposed with linear complexity by traversing the graph from leaves to root.

A depth-first search (DFS) starting from the (arbitrary) root is performed to find the correct processing order. The found nodes are stored in a list with the root as the last element and the last-found node as the first element. This list is then used in the modified LDU decomposition (Algorithm \ref{alg:factor}) and backsubstitution (Algorithm \ref{alg:solve}). \myupdate{Note that in the algorithms, vector indices $i$ stand for the respective rows of node $i$, and matrix indices $i,j$ stand for the respective rows of node $i$ and columns of node $j$.}

\begin{algorithm}[!htb]
	\begin{algorithmic}[1]
		\caption{Sparse In-Place LDU Decomposition, Complexity $\mathcal{O}(n)$}\label{alg:factor}
		\For{$i \in \text{list}$}\Comment{list from DFS}
		\For{$c \in \text{children}(i)$}\Comment{children from DFS}
		\State $\bs{F}_{i,c} \leftarrow \bs{F}_{i,c}\bs{F}_{c,c}^{-1}$
		\State $\bs{F}_{c,i} \leftarrow \bs{F}_{c,c}^{-1}\bs{F}_{c,i}$
		\State $\bs{F}_{i,i} \leftarrow \bs{F}_{i,i} - \bs{F}_{i,c}\bs{F}_{c,c}\bs{F}_{c,i}$
		\EndFor
		\EndFor
	\end{algorithmic}
\end{algorithm}

\begin{algorithm}[!htb]
	\begin{algorithmic}[1]
		\caption{Sparse In-Place LDU Backsubstitution, Complexity $\mathcal{O}(n)$}\label{alg:solve}
		\For{$i \in \text{list}$}
		\State $\Delta\bno{s}{i} \leftarrow -\bno{f}{i}$
		\For{$c \in \text{children}(i)$}
		\State $\Delta\bno{s}{i} \leftarrow \Delta\bno{s}{i} - \bs{F}_{i,c}\Delta\bno{s}{c}$
		\EndFor
		\EndFor
		\For{$i \in \text{reverse}(\text{list})$}
		\State $\Delta\bno{s}{i} \leftarrow \bs{F}_{i,i}^{-1}\Delta\bno{s}{i}$
		\State $\Delta\bno{s}{i} \leftarrow \Delta\bno{s}{i} - \bs{F}_{c,\mathrm{parent}(i)}\Delta\bno{s}{\mathrm{parent}(i)}$\Comment{ignore if $\mathrm{parent}(i)=\emptyset$}
		\EndFor
	\end{algorithmic}
\end{algorithm} 

The decomposition in Algorithm \ref{alg:factor} processes the matrix $\bs{F}$ according to the graph structure from leaves to root. Additionally, for each node, computations are only performed for connected components since computations for unconnected components are zero in the LDU decomposition. The linear complexity is a direct result. 

\textbf{Decomposition Complexity:} In an acyclic graph with $n$ nodes, each node has at most one parent, so there are $\mathcal{O}(n)$ children (and $\mathcal{O}(n)$ parents). Therefore, a total of $\mathcal{O}(n)$ evaluations of the for-loop on line 2 of Algorithm \ref{alg:factor} is required. The result is a linear complexity $\mathcal{O}(n)$.

\textbf{Backsubstitution Complexity:} In an acyclic graph with $n$ nodes there are $\mathcal{O}(n)$ children and $\mathcal{O}(n)$ parents. Therefore, a total of $\mathcal{O}(n)$ evaluations of the for-loops on lines 3 and 9 of Algorithm \ref{alg:solve} is required. The result is a linear complexity $\mathcal{O}(n)$.

\paragraph*{Articulated Mechanism Example}

The comparison of dense and sparse LDU decomposition for the example in Fig. \ref{fig:mechanism} is shown in Fig. \ref{fig:sparsity}.
\begin{figure}[!htb]
	\centering
	\begin{tikzpicture}
    \matrix [matrix of math nodes,
    nodes={rectangle, 
        minimum size=1.2em, text depth=0.25ex,
        inner sep=0pt, outer sep=0pt,
        anchor=center},
    column sep=-0.5\pgflinewidth,
    row sep=-0.5\pgflinewidth,
    inner sep=0pt,
    left delimiter=(, right delimiter=), 
    ] (m1)
    {
        \bs{D}_1 & 0 & 0 & 0 & 0 & \bs{c}'_{61} & 0 & 0 & \bs{c}'_{91}\\
        0 & \bs{D}_2 & 0 & 0 & 0 & \bs{c}'_{62} & \bs{c}'_{72} & \bs{c}'_{82} & 0\\
        0 & 0 & \bs{D}_3 & 0 & 0 & 0 & \bs{c}'_{73} & 0 & 0\\
        0 & 0 & 0 & \bs{D}_4 & 0 & 0 & 0 & \bs{c}'_{84} & 0\\
        0 & 0 & 0 & 0 & \bs{D}_5 & 0 & 0 & 0 & \bs{c}'_{95}\\
        \bs{c}_{61} & \bs{c}_{62} & 0 & 0 & 0 & \bs{D}_6 & \bullet & \bullet & \bullet\\
        0 & \bs{c}_{72} & \bs{c}_{73} & 0 & 0 & \bullet & \bs{D}_7 & \bullet & \bullet\\
        0 & \bs{c}_{82} & 0 & \bs{c}_{84} & 0 & \bullet & \bullet & \bs{D}_8 & \bullet\\
        \bs{c}_{91} & 0 & 0 & 0 & \bs{c}_{95} & \bullet & \bullet & \bullet & \bs{D}_9\\
    };

    \begin{scope}[on background layer]
    \fill[fill=red!30, rounded corners] 
    (m1-4-9.south west) -- (m1-1-9.south west) -- (m1-1-1.south west) -- (m1-1-1.north west) -- (m1-1-9.north east) -- (m1-4-9.south east);
    
    \draw[draw=black!50, rounded corners] 
    (m1-1-6.south west) -- (m1-1-1.south west) -- (m1-1-1.north west) -- (m1-1-9.north east) -- (m1-5-9.east);
    
    \draw[draw=black!50, rounded corners] 
    (m1-1-6.south east)  -- (m1-1-9.south west) -- (m1-4-9.south west);
    
    \fill[fill=yellow!30, rounded corners] 
    (m1-3-8.south west) -- (m1-2-8.south west) -- (m1-2-2.south west) -- (m1-2-2.north west) -- (m1-2-8.north east) -- (m1-3-8.south east);
    
    \draw[draw=black!50, rounded corners] 
    (m1-2-7.south west) -- (m1-2-2.south west) -- (m1-2-2.north west) -- (m1-2-6.north west);
    
    \draw[draw=black!50, rounded corners] 
    (m1-2-6.north east) -- (m1-2-8.north east) -- (m1-4-8.south east);
    
    \draw[draw=black!50, rounded corners] 
    (m1-2-8.south west)  -- (m1-4-8.south west);

    \filldraw[fill=yellow!30, draw=black!50,rounded corners] 
    (m1-5-9.north west) -- (m1-5-5.north west) -- (m1-5-5.south west) -- (m1-5-9.south east) 
    -- (m1-5-9.north east) ;
    
    \filldraw[fill=green!30, draw=black!50,rounded corners] 
    (m1-4-8.north west) -- (m1-4-4.north west) -- (m1-4-4.south west) -- (m1-4-8.south east) 
    -- (m1-4-8.north east) ;
    \filldraw[fill=green!30, draw=black!50,rounded corners] 
    (m1-3-7.north west) -- (m1-3-3.north west) -- (m1-3-3.south west) -- (m1-3-7.south east) 
    -- (m1-3-7.north east) ;
    \end{scope}
    
    \coordinate (rm1) at (5.0,0.0);

    \matrix [right of=rm1,matrix of math nodes,
    nodes={rectangle, 
        minimum size=1.2em, text depth=0.25ex,
        inner sep=0pt, outer sep=0pt,
        anchor=center},
    column sep=-0.5\pgflinewidth,
    row sep=-0.5\pgflinewidth,
    inner sep=0pt,
    left delimiter=(, right delimiter=),
    ] (m2)
    {
        \bs{D}_5 & \bs{c}'_{95} & 0 & 0 & 0 & 0 & 0 & 0 & 0 \\
        \bs{c}_{95} & \bs{D}_9 & 0 & 0 & 0 & 0 & 0 & 0 & \bs{c}_{91} \\	
        0 & 0 & \bs{D}_4 & \bs{c}'_{84} & 0 & 0 & 0 & 0 & 0 \\
        0 & 0 & \bs{c}_{84} & \bs{D}_8 & 0 & 0 &	\bs{c}_{82} & 0 & 0 \\
        0 & 0 & 0 & 0 & \bs{D}_3 & \bs{c}'_{73} & 0 & 0 & 0 \\
        0 & 0 & 0 & 0 & \bs{c}_{73} & \bs{D}_7 & \bs{c}_{72} & 0 & 0 \\
        0 & 0 & 0 & \bs{c}'_{82} & 0 & \bs{c}'_{72} & \bs{D}_2 & \bs{c}'_{62} & 0 \\
        0 & 0 & 0 & 0 & 0 & 0 & \bs{c}_{62} & \bs{D}_6 & \bs{c}_{61} \\	
        0 & \bs{c}'_{91} & 0 & 0 & 0 & 0 & 0 & \bs{c}'_{61} & \bs{D}_1\\
    };

    \begin{scope}[on background layer]
    \fill[fill=red!30, rounded corners] 
    (m2-2-2.north west) --
    (m2-9-2.south west) -- (m2-9-9.south east) -- (m2-9-9.north east) --
    (m2-9-2.north east) -- (m2-2-2.north east);
    \fill[fill=red!30, rounded corners] 
    (m2-8-8.north west) -- (m2-9-8.north west) -- (m2-9-7.north west) -- 
    (m2-9-7.south west) -- (m2-9-9.south east) -- (m2-9-9.north east) -- 
    (m2-9-8.north east) -- (m2-8-8.north east);
    
    \draw[draw=black!50, rounded corners] 
    (m2-2-2.north west) --
    (m2-9-2.south west) -- (m2-9-9.south east) -- (m2-9-9.north east) -- (m2-9-8.north east)
    -- (m2-8-8.north east);
    
    \draw[draw=black!50, rounded corners] 
    (m2-8-8.north west) --
    (m2-9-8.north west) -- (m2-9-2.north east) -- (m2-2-2.north east);
    
    \filldraw[fill=yellow!30,draw=black!50, rounded corners] 
    (m2-1-2.south east) --  (m2-1-2.north east) --(m2-1-1.north west) -- (m2-1-1.south west)
    -- (m2-1-2.south west);
    
    \fill[fill=yellow!30, rounded corners] 
    (m2-4-4.north west) -- (m2-7-4.south west) -- (m2-7-5.south east) -- (m2-7-5.north east) --
    (m2-7-4.north east) -- (m2-4-4.north east);
    \fill[fill=yellow!30, rounded corners] 
    (m2-6-6.north west) -- (m2-7-6.north west) -- (m2-7-5.north west) -- (m2-7-5.south west) -- (m2-7-7.south east) -- (m2-7-7.north east) -- 
    (m2-7-6.north east) -- (m2-6-6.north east);
    \fill[fill=yellow!30, rounded corners] 
    (m2-7-8.south west) -- (m2-7-5.south west) -- (m2-7-5.north west) -- (m2-7-8.north east) -- 
    (m2-7-8.south east);
    
    \draw[draw=black!50, rounded corners] 
    (m2-4-4.north west) --
    (m2-7-4.south west) -- (m2-7-8.south west);
    
    \draw[draw=black!50, rounded corners] 
    (m2-7-8.south east) -- (m2-7-8.north east) -- (m2-7-6.north east)
    -- (m2-6-6.north east);
    
    \draw[draw=black!50, rounded corners] 
    (m2-6-6.north west) --
    (m2-7-6.north west) -- (m2-7-4.north east) -- (m2-4-4.north east);
    
    \filldraw[fill=green!30,draw=black!50, rounded corners] 
    (m2-3-4.south east) --  (m2-3-4.north east) --(m2-3-3.north west) -- (m2-3-3.south west)
    -- (m2-3-4.south west);
    
    \filldraw[fill=green!30,draw=black!50, rounded corners] 
    (m2-5-6.south east) --  (m2-5-6.north east) --(m2-5-5.north west) -- (m2-5-5.south west)
    -- (m2-5-6.south west);
    \end{scope}

    \node at ($(m1-1-1)-(0.8,0)$) {(a)};
    \node at ($(m2-1-1)-(0.8,0)$) {(b)};
    \end{tikzpicture}
	\caption{Matrices for the mechanism in Fig. \ref{fig:mechanism} with matching color scheme. Fill-in indicated with ``$\bullet$''. (a) Unordered matrix with fill-in after LDU decomposition. (b) Rearranged matrix without fill-in after LDU decomposition.
	}
	\label{fig:sparsity}
\end{figure}

The matrices in Fig. \ref{fig:sparsity} have off-diagonal entries only at the intersection of directly connected nodes. For the example mechanism in Fig. \ref{fig:mechanism}, the diagonal entries $\bs{D}_1$ to $\bs{D}_5$ are the derivatives of the dynamics $\bs{d}$ of each body, and the off-diagonal entries $\bs{c}_{ij}$ are the equality constraint derivatives representing joints between two bodies. Note that when processing the matrix in the wrong order, so-called fill-in is created. This fill-in occurs at the off-diagonals of nodes indirectly connected through a node that is processed before it becomes a leaf. Since fill-in must also be processed in the decomposition and backsubstitution, linear complexity is no longer achieved.

\begin{figure}[!htb]
	\centering
	\begin{tikzpicture}

    \coordinate (Ca) at (0.0,0.0);	
    \node at ($(Ca)+(-1.5,0.0)$) {(a)};
    
    \draw[fill=black!5,rounded corners=6pt,rotate=-75]
    ($(Ca)+(0,0.0)$) rectangle ++(2,-0.4);
    \draw[fill=black!5,rounded corners=6pt,rotate=-50]
    ($(Ca)+(-0.1,-0.07)$) rectangle ++(2,-0.4);
    \draw[fill=black!5,rounded corners=6pt,rotate=-140]
    ($(Ca)+(0.75,1.8)$) rectangle ++(2,-0.4);
    \draw[fill=black!5,rounded corners=6pt,rotate=-95]
    ($(Ca)+(1.22,1.22)$) rectangle ++(2,-0.4);
    
    \draw[fill=black] ($(Ca)+(0.27,-1.85)$) circle (0.065);
    \draw[fill=black] ($(Ca)+(-0.14,-0.25)$) circle (0.065);
    \draw[fill=black] ($(Ca)+(0.89,-1.52)$) circle (0.065);
    
    \node [shading = axis,rectangle, left color=darkgray!50!white, right color=white, shading angle=0, minimum width = 70] at ($(Ca)+(-0.1,-3.42)$) {};

    \node at ($(Ca)+(0.15,-1.35)$) {1};
    \node at ($(Ca)+(-0.4,-2.4)$) {2};
    \node at ($(Ca)+(0.35,-0.8)$) {3};
    \node at ($(Ca)+(0.8,-2.4)$) {4};
    
    \draw ($(Ca)+(0.27,-1.85)$) -- ($(Ca)+(-0.35,-1.6)$);
    \draw ($(Ca)+(-0.14,-0.25)$) -- ($(Ca)+(-0.65,-0.2)$);
    \draw ($(Ca)+(0.89,-1.52)$) -- ($(Ca)+(1.0,-0.65)$);
    
    \node at ($(Ca)+(-0.5,-1.6)$){5};
    \node at ($(Ca)+(-0.8,-0.2)$) {6};
    \node at ($(Ca)+(1.0,-0.5)$) {7};

    \node at (2,0) {(b)};
    
    \draw (2.8,-0.2) -- (2.25,-0.95);
    \draw (2.8,-0.2) -- (3.35,-0.95);
    \draw  (2.25,-0.95) -- (2.25,-1.8);
    \draw (3.35,-0.95) -- (3.35,-1.8);
    \draw (2.25,-1.8) -- (2.25,-2.65);
    \draw (3.35,-1.8) -- (3.35,-2.65);
    \draw (2.25,-2.65) -- (2.25,-3.5);
    \draw (3.35,-2.65) -- (3.35,-3.5);

    \draw[fill=red!30] (2.8,-0.2) circle (0.275);
    \draw[fill=red!30] (2.0,-0.7) rectangle ++(0.5,-0.5);
    \draw[fill=red!30] (3.1,-0.7) rectangle ++(0.5,-0.5);
    \draw[fill=yellow!30] (2.25,-1.8) circle (0.275);
    \draw[fill=yellow!30] (3.35,-1.8) circle (0.275);
    \draw[fill=yellow!30] (2.0,-2.4) rectangle ++(0.5,-0.5);
    \draw[fill=yellow!30] (3.1,-2.4) rectangle ++(0.5,-0.5);
    \node[draw=black, fill=green!30, regular polygon,regular polygon sides=3] at (2.25,-3.5) {};
    \node[draw=black, fill=green!30, regular polygon,regular polygon sides=3] at (3.35,-3.5) {};
    
    \node at (2.25,-0.95) {1};
    \node at (2.25,-2.65) {2};
    \node at (3.35,-0.95) {3};
    \node at (3.35,-2.65) {4};
    \node at (2.25,-1.8) {5};
    \node at (2.8,-0.2) {6};
    \node at (3.35,-1.8) {7};
    \node at (2.25,-3.5) {8};
    \node at (3.35,-3.5) {9};

    \node at (4.35,0) {(c)};

    \coordinate (rm1) at (6.0,-1.75);

    \matrix [right of=rm1,matrix of math nodes,
    nodes={rectangle, 
        minimum size=1.2em, text depth=0.25ex,
        inner sep=0pt, outer sep=0pt,
        anchor=center},
    column sep=-0.5\pgflinewidth,
    row sep=-0.5\pgflinewidth,
    inner sep=0pt,
    left delimiter=(, right delimiter=),
    ] (m2)
    {
        \bs{D}_9 & \bs{c}_{94} & 0 & 0 & 0 & 0 & 0 & 0 & 0 \\
        \bs{c}'_{94} & \bs{D}_4 & \bs{c}'_{74} & 0 & 0 & 0 & 0 & 0 & 0 \\	
        0 & \bs{c}_{74} & \bs{D}_7 & \bs{c}_{73} & 0 & 0 & 0 & 0 & 0 \\
        0 & 0 & \bs{c}'_{73} & \bs{D}_3 & \bs{c}'_{63} & 0 & 0 & 0 & 0 \\
        0 & 0 & 0 & \bs{c}_{63} & \bs{D}_6 & 0 & 0 & 0 & \bs{c}_{61} \\
        0 & 0 & 0 & 0 & 0 & \bs{D}_8 & \bs{c}_{82} & 0 & 0 \\
        0 & 0 & 0 & 0 & 0 & \bs{c}'_{82} & \bs{D}_2 & \bs{c}'_{52} & 0 \\
        0 & 0 & 0 & 0 & 0 & 0 & \bs{c}_{52} & \bs{D}_5 & \bs{c}_{51} \\	
        0 & 0 & 0 & 0 & \bs{c}'_{61} & 0 & 0 & \bs{c}'_{51} & \bs{D}_1\\
    };

    \begin{scope}[on background layer]
    \fill[fill=red!30, rounded corners]
    (m2-4-3.north west) -- (m2-4-3.south west) -- (m2-4-5.south west) -- (m2-9-5.south west) -- (m2-9-9.south east) -- (m2-9-9.north east) -- (m2-9-5.north east) -- (m2-4-5.north east) -- (m2-4-3.north east);

    \fill[fill=yellow!30, rounded corners]
    (m2-8-8.south west) -- (m2-7-8.south west) -- (m2-7-7.south west) -- (m2-7-7.north west) -- (m2-7-8.north east) -- (m2-8-8.south east); 
    \fill[fill=yellow!30, rounded corners]
    (m2-7-6.north west) -- (m2-7-6.south west) -- (m2-7-7.south east) -- (m2-7-7.north east) -- (m2-7-6.north east); 

    \fill[fill=yellow!30, rounded corners]
    (m2-3-3.south west) -- (m2-2-3.south west) -- (m2-2-2.south west) -- (m2-2-2.north west) -- (m2-2-3.north east) -- (m2-3-3.south east); 
    \fill[fill=yellow!30, rounded corners]
    (m2-2-1.north west) -- (m2-2-1.south west) -- (m2-2-2.south east) -- (m2-2-2.north east) -- (m2-2-1.north east); 

    \filldraw[fill=green!30,draw=black!50, rounded corners] 
    (m2-6-6.south west) -- (m2-6-6.north west) -- (m2-6-6.north east) -- (m2-6-6.south east);

    \filldraw[fill=green!30,draw=black!50, rounded corners] 
    (m2-1-1.south west) -- (m2-1-1.north west) -- (m2-1-1.north east) -- (m2-1-1.south east);

    \draw[draw=black!50, rounded corners] 
    (m2-9-9.north west) -- (m2-9-9.north east) -- (m2-9-9.south east) -- (m2-9-5.south west) -- (m2-4-5.south west) -- (m2-4-3.south west) -- (m2-4-3.north west);
    \draw[draw=black!50, rounded corners] 
    (m2-9-8.north west) -- (m2-9-5.north east) -- (m2-4-5.north east) -- (m2-4-3.north east);
    \draw[draw=black!50, rounded corners] 
    (m2-8-8.south east) -- (m2-7-8.north east) -- (m2-7-7.north west);
    \draw[draw=black!50, rounded corners] 
    (m2-8-8.south west) -- (m2-7-8.south west) -- (m2-7-6.south west) -- (m2-7-6.north west);
    \draw[draw=black!50, rounded corners] 
    (m2-3-3.south east) -- (m2-2-3.north east) -- (m2-2-2.north west);
    \draw[draw=black!50, rounded corners] 
    (m2-3-3.south west) -- (m2-2-3.south west) -- (m2-2-1.south west) -- (m2-2-1.north west);

    \end{scope} 
    \end{tikzpicture}
	\caption{(a) A walking mechanism with four links, three joints, and two contact points. (b) A graph representing the mechanism. (c) The corresponding rearranged matrix without fill-in after decomposition.}
	\label{fig:mechanism2}
\end{figure}
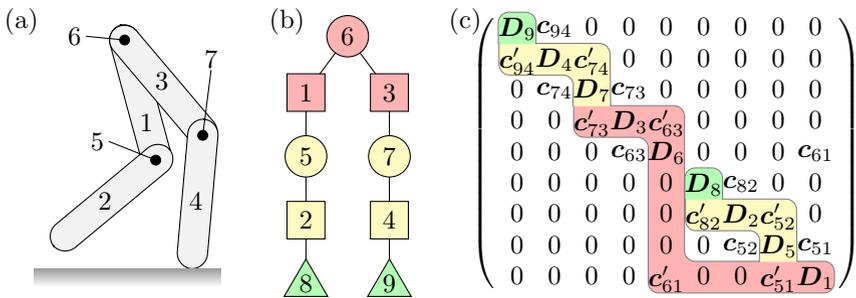

\paragraph*{Environment Contact Example}
A direct result of the linear-complexity property for acyclic graphs is the following. Mechanisms without kinematic loops and only environment contact, i.e., no contact between bodies, correspond to acyclic graphs and, therefore, have linear complexity in the number of bodies and contact points. Consider the exemplary mechanism, graph, and matrix in Fig. \ref{fig:mechanism2}.

Since all environment contacts are leaves in the graph, adding contact points leads to linear scaling with the correct processing order. This result is especially interesting for bipedal or quadrupedal walking robots.

\subsection{Reduced-Fill-In Algorithm}
\begin{figure}[!htb]
	\centering
	\begin{tikzpicture}

    \coordinate (Ca) at (0.0,0.0);	
    \node at ($(Ca)+(-1.5,0.0)$) {(a)};
    
    \draw[fill=black!5,rounded corners=6pt]
    ($(Ca)+(0,0.0)$) rectangle ++(2,-0.4);
    \draw[fill=black!5,rounded corners=6pt,rotate=-45]
    ($(Ca)+(0.2,0.325)$) rectangle ++(0.4,-1.8);
    \draw[fill=black!5,rounded corners=6pt,rotate=10]
    ($(Ca)+(-1.015,-0.86)$) rectangle ++(0.4,-1.4);
    \draw[fill=black!5,rounded corners=6pt,rotate=15]
    ($(Ca)+(1.3,-0.4)$) rectangle ++(0.4,-1.4);
    \draw[fill=black!5,rounded corners=6pt,rotate=80]
    ($(Ca)+(-1.01,0.27)$) rectangle ++(0.4,-2.5);
    
    \draw[fill=black] ($(Ca)+(0.35,-0.2)$) circle (0.065);
    \draw[fill=black] ($(Ca)+(-0.62,-1.17)$) circle (0.065);
    \draw[fill=black] ($(Ca)+(-0.25,-0.8)$) circle (0.065);
    \draw[fill=black] ($(Ca)+(1.6,-0.2)$) circle (0.065);
    \draw[fill=black] ($(Ca)+(1.85,-1.15)$) circle (0.065);
    
    \node at ($(Ca)+(1,-0.2)$) {1};
    \node at ($(Ca)+(0.1,-0.47)$) {2};
    \node at ($(Ca)+(-0.55,-1.7)$) {3};
    \node at ($(Ca)+(0.8,-0.95)$) {4};
    \node at ($(Ca)+(1.75,-0.7)$) {5};
    
    \draw ($(Ca)+(0.35,-0.2)$) -- ($(Ca)+(0.8,0.2)$);
    \draw ($(Ca)+(-0.62,-1.17)$) -- ($(Ca)+(-0.92,-0.85)$);
    \draw ($(Ca)+(-0.25,-0.8)$) -- ($(Ca)+(-0.55,-0.55)$);
    \draw ($(Ca)+(1.6,-0.2)$) -- ($(Ca)+(1.2,0.2)$);
    \draw ($(Ca)+(1.85,-1.15)$) -- ($(Ca)+(2.17,-1.47)$);
    
    \node at ($(Ca)+(1.0,0.2)$) {6};
    \node at ($(Ca)+(-1.05,-0.75)$) {7};
    \node at ($(Ca)+(-0.7,-0.45)$) {8};
    \node at ($(Ca)+(2.32,-1.5)$) {9};

    \node at (4,0) {(b)};
    
    \draw (6.2,-0.2) -- (5.5,-0.5);
    \draw (5.5,-0.45) -- (5.1,-1.2);
    \draw (5.3,-1.1) -- (4.5,-1.4);
    \draw (4.5,-1.4) -- (4.3,-2);
    \draw (5.2,-1.1) -- (6.1,-1.4);
    \draw (6.2,-1.4) -- (6.5,-1.9);
    \draw (5.5,-0.5) -- (7.9,-1.3);
    \draw (7.1,-1.6) -- (7.9,-1.2);
    \draw (7.1,-1.6) -- (6.5,-2.2);

    \draw[fill=red!30] (6,0.0) rectangle ++(0.5,-0.5);
    \draw[fill=yellow!30] (5,-0.9) rectangle ++(0.5,-0.5);
    \draw[fill=green!30] (4.1,-1.8) rectangle ++(0.5,-0.5);
    \draw[fill=green!30] (6.3,-1.8) rectangle ++(0.5,-0.5);
    \draw[fill=yellow!30] (7.5,-1.1) rectangle ++(0.5,-0.5);
    
    \draw[fill=red!30] (5.5,-0.45) circle (0.275);
    \draw[fill=yellow!30] (4.5,-1.4) circle (0.275);
    \draw[fill=yellow!30] (6.1,-1.4) circle (0.275);
    \draw[fill=yellow!30] (7.1,-1.6) circle (0.275);
    
    \node at (6.25,-0.25) {1};
    \node at (5.25,-1.15) {2};
    \node at (4.35,-2.05) {3};
    \node at (6.55,-2.05) {4};
    \node at (7.75,-1.35) {5};
    \node at (5.5,-0.45) {6};
    \node at (4.5,-1.4) {7};
    \node at (6.1,-1.4) {8};
    \node at (7.1,-1.6) {9};
    \end{tikzpicture}
	\caption{(a) A mechanism with five links and four joints containing a kinematic loop. (b) A cyclic graph representing the mechanism and its matrix.}
	\label{fig:mechanism3}
\end{figure}
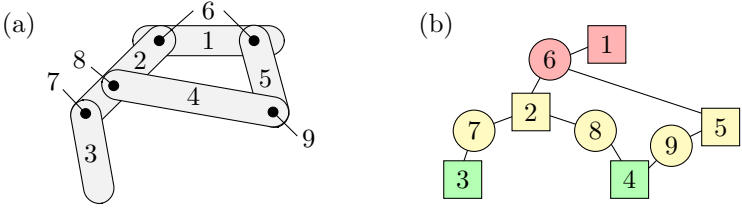

If the graph of a mechanism has cycles, fill-in can generally no longer be avoided entirely since nodes of a cycle are never leaves. However, by processing the leaves attached to a cycle first, the amount of fill-in is reduced. As an example, the mechanism from Fig. \ref{fig:mechanism} is modified to contain a kinematic loop, displayed in Fig. \ref{fig:mechanism3}.

Note that for computational reasons, two joints are combined into node 6 since these two joints form the beginning of the kinematic loop. Such loop-openers are also found with the depth-first search since they are simply the first and last joint in a detected loop. In the algorithms, a distinction is made between nodes that are part of a cycle and nodes that are not. This distinction is also determined by the depth-first search.

Since any square matrix can be represented by a (potentially cyclic) graph, the following algorithms are applicable to all such matrices. In the case of mechanical systems, they can be used for all components defined in Section \ref{sec:simulator_components}. The sparse decomposition for such systems is formulated in Algorithm \ref{alg:factor2} and the backsubstitution in Algorithm \ref{alg:solve2}.

\begin{algorithm}[!htb]
	\begin{algorithmic}[1]
		\caption{Sparse In-Place LDU Decomposition, Complexity $\mathcal{O}(n+kn^2)$}\label{alg:factor2}
		\For{$i \in \mathrm{list}$}\Comment{list from DFS}
			\For{$c_1 \in \mathrm{acyclic\_children}(i)$}\Comment{children not in a cycle}
				\State $\bs{F}_{i,c_1} \leftarrow \bs{F}_{i,c_1}\bs{F}_{c_1,c_1}^{-1}$
				\State $\bs{F}_{c_1,i} \leftarrow \bs{F}_{c_1,c_1}^{-1}\bs{F}_{c_1,i}$
				\State $\bs{F}_{i,i} \leftarrow \bs{F}_{i,i} - \bs{F}_{i,c_1}\bs{F}_{c_1,c_1}\bs{F}_{c_1,i}$
			\EndFor
			\For{$c_1 \in \mathrm{cyclic\_children}(i)$}\Comment{successors of $i$ in a cycle started at $i$}
				\For{$c_2 \in \mathrm{cyclic\_children}(i)$}
					\If{$c_1 == c_2$}
						\State{break}
					\ElsIf{$c_2 \notin \mathrm{all\_children}(c_1)$}\Comment{acyclic and cyclic children}
						\State{continue}
					\Else
						\State $\bs{F}_{i,c_1} \leftarrow \bs{F}_{i,c_1} - \bs{F}_{i,c_2}\bs{F}_{c_2,c_2}\bs{F}_{c_2,c_1}$
						\State $\bs{F}_{c_1,i} \leftarrow \bs{F}_{c_1,i} - \bs{F}_{c_1,c_2}\bs{F}_{c_2,c_2}\bs{F}_{c_2,i}$
					\EndIf
				\EndFor
				\State $\bs{F}_{i,c_1} \leftarrow \bs{F}_{i,c_1}\bs{F}_{c_1,c_1}^{-1}$
				\State $\bs{F}_{c_1,i} \leftarrow \bs{F}_{c_1,c_1}^{-1}\bs{F}_{c_1,i}$
				\State $\bs{F}_{i,i} \leftarrow \bs{F}_{i,i} - \bs{F}_{i,c_1}\bs{F}_{c_1,c_1}\bs{F}_{c_1,i}$
			\EndFor
		\EndFor
	\end{algorithmic}
\end{algorithm}

\begin{algorithm} 
	\begin{algorithmic}[1]
		\caption{Sparse In-Place LDU Backsubstitution, Complexity $\mathcal{O}(n+nk)$}\label{alg:solve2}
		\For{$i \in \mathrm{list}$}
			\State $\Delta\bno{s}{i} \leftarrow -\bno{f}{i}$
			\For{$c \in  \mathrm{all\_children}(i)$}
				\State $\Delta\bno{s}{i} \leftarrow \Delta\bno{s}{i} - \bs{F}_{i,c}\Delta\bno{s}{c}$
			\EndFor
		\EndFor
		\For{$i \in \mathrm{reverse}(\mathrm{list})$}
			\State $\Delta\bno{s}{i} \leftarrow \bs{F}_{i,i}^{-1}\Delta\bno{s}{i}$
			\For{$p \in \mathrm{all\_parents}(i)$}\Comment{direct and loop-opening parents}
				\State $\Delta\bno{s}{i} \leftarrow \Delta\bno{s}{i} - \bs{F}_{c,p}\Delta\bno{s}{p}$ 
			\EndFor
		\EndFor
	\end{algorithmic}
\end{algorithm}

Note that all lists in Algorithms \ref{alg:factor2} and \ref{alg:solve2} are sorted in the same order as the depth-first search list. 

\textbf{Decomposition Complexity:} In a cyclic graph with $n$ nodes and $k$ cycles, there are $\mathcal{O}(n)$ acyclic children and $\mathcal{O}(n)$ acyclic parents. Additionally, there are $\mathcal{O}(n)$ cyclic children and $\mathcal{O}(n)$ loop-opening parents per cycle since each cycle contains at most all nodes, and each of these nodes has the same loop-opening parent. Therefore, a total of $\mathcal{O}(n)$ evaluations of the for-loop on line 2 of Algorithm \ref{alg:factor2} and a total of $\mathcal{O}(kn^2)$ evaluations of the for-loop on line 8 is required. Resulting is a complexity $\mathcal{O}(n+kn^2)$---quadratic in $n$ and linear in $k$.

\textbf{Backsubstitution Complexity:} In a cyclic graph with $n$ nodes and $k$ cycles, there are $\mathcal{O}(n)$ acyclic children, $\mathcal{O}(n)$ acyclic parents, $\mathcal{O}(n)$ cyclic children per cycle, and $\mathcal{O}(n)$ loop-opening parents per cycle. Therefore, a total of $\mathcal{O}(n+nk)$ evaluations of the for-loop on line 3 of Algorithm \ref{alg:solve2} and a total of $\mathcal{O}(n+nk)$ evaluations of the for-loop on line 9 is required. Resulting is a complexity $\mathcal{O}(n+nk)$---linear in $n$ and linear in $k$.

In the worst case of a fully connected graph, i.e., a fully dense matrix, each node has $\mathcal{O}(n)$ cyclic children and parents, resulting in a complexity of $\mathcal{O}(n+n^3)$. However, for real systems, the theoretical complexities are often too conservative. For a system without intersecting cycles, i.e., each node belongs to at most a single cycle, there is a total of $\mathcal{O}(n)$ cyclic children and parents and, therefore, the overall complexity is $\mathcal{O}(n+n^2)$. In the case of a constant cycle size, there is a total of $\mathcal{O}(k)$ cyclic children and parents, and a linear complexity $\mathcal{O}(n + k)$ is obtained. In a combined setting with non-intersecting cycles of fixed size, for example, a mechanism with disconnected identical legs made of kinematic loops, there are $\mathcal{O}(n)$ cycles with $\mathcal{O}(1)$ cyclic children and parents each, resulting in a linear complexity $\mathcal{O}(n)$. Improvements in the complexity are theoretically possible, but finding a processing order that creates the minimum fill-in is generally NP-hard \cite{duff_direct_2017}.

The comparison of dense and sparse decomposition for the example in Fig. \ref{fig:mechanism2} with additional dampers at each joint is shown in Fig. \ref{fig:sparsity2}.

\begin{figure}[!htb]
	\centering
	\begin{tikzpicture}
    \matrix [matrix of math nodes,
    nodes={rectangle, 
        minimum size=1.2em, text depth=0.25ex,
        inner sep=0pt, outer sep=0pt,
        anchor=center},
    column sep=-0.5\pgflinewidth,
    row sep=-0.5\pgflinewidth,
    inner sep=0pt,
    left delimiter=(, right delimiter=), 
    ] (m1)
    {
        \bs{D}_1 & \bs{c}'_{12} & 0 & 0 & \bs{c}'_{51} & \bs{c}'_{61} & 0 & 0 & 0\\
        \bs{c}_{12} & \bs{D}_2 & \bs{c}'_{32} & \bs{c}'_{42} & \bullet & \bs{c}'_{62} & \bs{c}'_{72} & \bs{c}'_{82} & 0\\
        0 & \bs{c}_{32} & \bs{D}_3 & \bullet & \bullet & \bullet & \bs{c}'_{73} & \bullet & 0\\
        0 & \bs{c}_{42} & \bullet & \bs{D}_4 & \bs{c}'_{54} & \bullet & \bullet & \bs{c}'_{84} & \bs{c}'_{94}\\
        \bs{c}_{51} & \bullet & \bullet & \bs{c}_{54} & \bs{D}_5 & \bs{c}'_{65} & \bullet & \bullet & \bs{c}'_{95}\\
        \bs{c}_{61} & \bs{c}_{62} & \bullet & \bullet & \bs{c}_{65} & \bs{D}_6 & \bullet & \bullet & \bullet\\
        0 & \bs{c}_{72} & \bs{c}_{73} & \bullet & \bullet & \bullet & \bs{D}_7 & \bullet & \bullet\\
        0 & \bs{c}_{82} & \bullet & \bs{c}_{84} & \bullet & \bullet & \bullet & \bs{D}_8 & \bullet\\
        0 & 0 & 0 & \bs{c}_{94} & \bs{c}_{95} & \bullet & \bullet & \bullet & \bs{D}_9\\
    };

    \begin{scope}[on background layer]
    \filldraw[fill=red!30, draw=black!50, rounded corners] 
    (m1-1-6.south west) -- (m1-1-1.south west) -- (m1-1-1.north west) -- (m1-1-6.north east) -- (m1-1-6.south east);
    
    \fill[fill=yellow!30, rounded corners] 
    (m1-3-8.south west) -- (m1-2-8.south west) -- (m1-2-2.south west) -- (m1-2-2.north west) -- (m1-2-8.north east) -- (m1-3-8.south east);

    \draw[draw=black!50, rounded corners] 
    (m1-2-6.north east)  -- (m1-2-8.north east) -- (m1-3-8.south east);
    
    \draw[draw=black!50, rounded corners] 
    (m1-2-7.south west) -- (m1-2-2.south west) -- (m1-2-2.north west) -- (m1-2-6.north west);
    
    \filldraw[fill=green!30, draw=black!50,rounded corners] 
    (m1-3-7.north west) -- (m1-3-3.north west) -- (m1-3-3.south west) -- (m1-3-7.south east) 
    -- (m1-3-7.north east) ;
    
    \fill[fill=green!30, rounded corners] 
    (m1-4-9.south east) -- (m1-4-9.north east) -- (m1-4-4.north west) -- (m1-4-4.south west) -- (m1-4-9.south east);

    \draw[draw=black!50, rounded corners] 
    (m1-4-9.north west)  -- (m1-4-9.north east) -- (m1-4-9.south east);

    \draw[draw=black!50, rounded corners] 
    (m1-4-8.north west)  -- (m1-4-4.north west) -- (m1-4-4.south west) -- (m1-4-8.south east);

    \filldraw[fill=yellow!30, draw=black!50,rounded corners] 
    (m1-5-9.north west) -- (m1-5-5.north west) -- (m1-5-5.south west) -- (m1-5-9.south east) 
    -- (m1-5-9.north east) ;

    \draw[draw=black!50, dashed] 
    (m1-1-6.south west)  -- (m1-5-6.north west);

    \draw[draw=black!50, dashed] 
    (m1-1-6.south east)  -- (m1-5-6.north east);
\end{scope}
    
    \coordinate (rm1) at (5.0,0.0);

    \matrix [right of=rm1,matrix of math nodes,
    nodes={rectangle, 
        minimum size=1.2em, text depth=0.25ex,
        inner sep=0pt, outer sep=0pt,
        anchor=center},
    column sep=-0.5\pgflinewidth,
    row sep=-0.5\pgflinewidth,
    inner sep=0pt,
    left delimiter=(, right delimiter=),
    ] (m2)
    {
        \bs{D}_5 & \bs{c}'_{95} & \bs{c}_{54} & 0 & 0 & 0 & 0 & \bs{c}'_{65} & \bs{c}_{51} \\
        \bs{c}_{95} & \bs{D}_9 & \bs{c}_{94} & 0 & 0 & 0 & 0 & \bullet & 0 \\	
        \bs{c}'_{54} & \bs{c}'_{94} & \bs{D}_4 & \bs{c}'_{84} & 0 & 0 & \bs{c}_{42} & \bullet & 0 \\
        0 & 0 & \bs{c}_{84} & \bs{D}_8 & 0 & 0 & \bs{c}_{82} & \bullet & 0 \\
        0 & 0 & 0 & 0 & \bs{D}_3 & \bs{c}'_{73} & \bs{c}_{32} & 0 & 0 \\
        0 & 0 & 0 & 0 & \bs{c}_{73} & \bs{D}_7 & \bs{c}_{72} & 0 & 0 \\
        0 & 0 & \bs{c}'_{42} & \bs{c}'_{82} & \bs{c}'_{32} & \bs{c}'_{72} & \bs{D}_2 & \bs{c}'_{62} & \bs{c}_{21} \\
        \bs{c}_{65} & \bullet & \bullet & \bullet & 0 & 0 & \bs{c}_{62} & \bs{D}_6 & \bs{c}_{61} \\	
        \bs{c}'_{51} & 0 & 0 & 0 & 0 & 0 & \bs{c}'_{21} & \bs{c}'_{61} & \bs{D}_1\\
    };

    \begin{scope}[on background layer]
        \fill[fill=yellow!30, rounded corners] 
        (m2-4-4.north east) -- (m2-7-4.north east) -- (m2-7-6.north east) -- (m2-7-6.south east) -- (m2-7-4.south west) -- (m2-4-4.north west);
        \fill[fill=yellow!30, rounded corners] 
        (m2-6-6.north east) -- (m2-7-6.north east) -- (m2-7-7.north east) -- (m2-7-7.south east) -- (m2-7-5.south west) -- (m2-7-5.north west) -- (m2-7-6.north west) -- (m2-6-6.north west);
        \fill[fill=yellow!30, rounded corners] 
        (m2-7-8.south west) -- (m2-7-6.south west) -- (m2-7-6.north west) -- (m2-7-8.north west) -- (m2-1-8.south west) -- (m2-1-3.south west) -- (m2-1-3.north west) -- (m2-1-8.north east) -- (m2-7-8.south east);
        \fill[fill=yellow!30, rounded corners] 
        (m2-2-2.south west) -- (m2-1-2.south west) -- (m2-1-1.south west) -- (m2-1-1.north west) -- (m2-1-3.north east) -- (m2-1-3.south east) -- (m2-1-2.south east) -- (m2-2-2.south east);

        \fill[fill=green!30, rounded corners] 
        (m2-3-4.south west) -- (m2-3-3.south west) -- (m2-3-3.north west) -- (m2-3-4.north east) -- (m2-3-4.south east);
        \fill[fill=green!30, rounded corners] 
        (m2-3-3.north east) -- (m2-3-3.south east) -- (m2-3-3.south west) -- (m2-3-2.south west) -- (m2-3-2.north west);

        \filldraw[fill=red!30, draw=black!50, rounded corners] 
        (m2-8-8.north west) -- (m2-9-8.south west) -- (m2-9-9.south east) -- (m2-9-9.north east) -- (m2-8-8.south east) -- (m2-8-8.north east);
        
        \filldraw[fill=green!30, draw=black!50, rounded corners] 
        (m2-5-6.south west) -- (m2-5-5.south west) -- (m2-5-5.north west) -- (m2-5-6.north east) -- (m2-5-6.south east);

        \draw[draw=black!50, rounded corners] 
        (m2-7-8.south east) -- (m2-1-8.north east) -- (m2-1-1.north west) -- (m2-1-1.south west) -- (m2-1-2.south west) -- (m2-3-2.south west) -- (m2-3-3.south east);

        \draw[draw=black!50, rounded corners] 
        (m2-4-4.north west) -- (m2-7-4.south west) -- (m2-7-7.south east);

        \draw[draw=black!50, rounded corners] 
        (m2-3-3.north west) -- (m2-3-4.north east) -- (m2-7-4.north east) -- (m2-7-6.north west) -- (m2-6-6.north west);

        \draw[draw=black!50, rounded corners] 
        (m2-6-6.north east) -- (m2-7-6.north east) -- (m2-7-7.north east) -- (m2-1-8.south west) -- (m2-1-2.south east) -- (m2-2-2.south east);
    
    \end{scope}

    \node at ($(m1-1-1)-(0.8,0)$) {(a)};
    \node at ($(m2-1-1)-(0.8,0)$) {(b)};
    \end{tikzpicture}
	\caption{Matrices for the mechanism in Fig. \ref{fig:mechanism3}. (a) Unordered, almost fully dense matrix due to fill-in after LDU decomposition. (b) Rearranged matrix with minimal fill-in after LDU decomposition. 
	}
	\label{fig:sparsity2}
\end{figure}

The dampers depend on the relative velocity between the two connected bodies, leading to additional off-diagonal entries. As Fig. \ref{fig:sparsity2} shows, with a bad processing order, an almost fully dense matrix is obtained due to fill-in, while the correct processing order creates fill-in only at the off-diagonals of nodes that are part of cycles and the loop-openers.

\section{Evaluation}\label{sec:evaluation}
The evaluation of the simulator is comprised of \myupdate{four} parts. First, the physical accuracy of the variational integrator is analyzed. Then, the runtime and computational complexity of the graph-based algorithms are investigated. \myupdate{Next, the numerical robustness of the simulator is tested.} Lastly, two application examples for the simulator are given. 
Comparisons are made \myupdate{to different simulators and integrators.} We compare to the dynamics simulator RigidBodyDynamics \cite{koolen_julia_2019} as it is written in the same programming language as our integrator, Julia \cite{bezanson_julia_2017}, and to the widely used simulator MuJoCo which is representative for soft constraint handling. {Variational integrator comparisons are made with two minimal-coordinate integrators \cite{lee_linear-time_2016,fan_efficient_2018} and an integrator in redundant coordinates for constrained systems \cite{kinon_ggl_2023}.} 
In order to solve the integrator equations \eqref{eqn:complete_integrator}, a basic interior-point method is used and described in Appendix \ref{sec:app_simulator}. Note that the focus of the implementation is on the theoretical properties and not runtime optimization. Nonetheless, even this basic implementation achieves reasonable timing results, benefitting from the easy and modular implementation in maximal coordinates. Code for the simulator\footnote[1]{\url{https://github.com/janbruedigam/ConstrainedDynamics.jl}} and the graph-based system solver\footnote[2]{\url{https://github.com/janbruedigam/GraphBasedSystems.jl}} including all experiments and additional examples is available in Julia. All experiments are carried out on an Asus ZenBook with an i7-8565U CPU and 16GB RAM.

\subsection{Physical Accuracy}
\begin{figure}[!htb]
	\centering
	\input{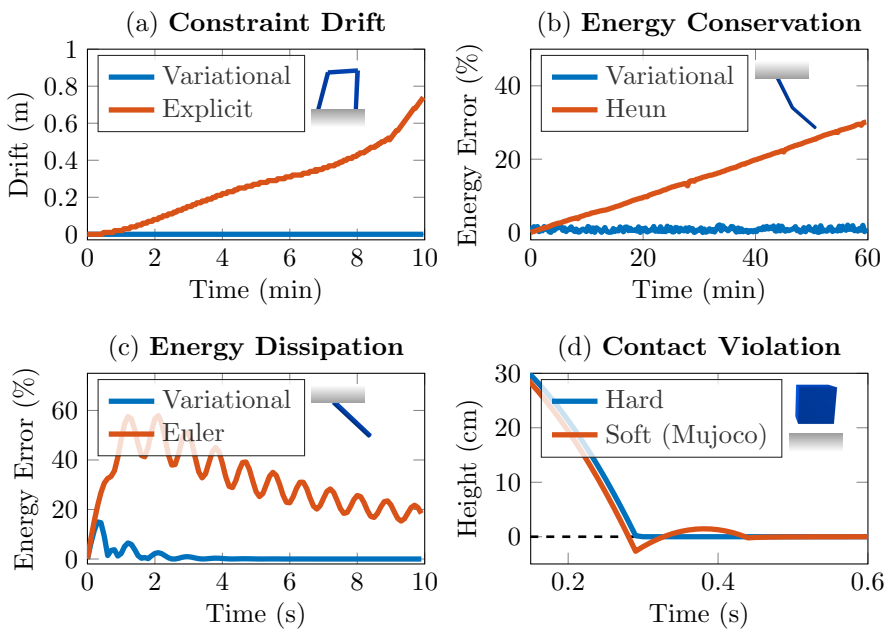}
	\caption{Evaluation of the physical accuracy of the variational integrator (blue) with comparison (red). (a) Zero constraint drift with variational integrator, drift resulting from explicit 4th-order Runge-Kutta. (b) Bounded energy error with variational integrator for conservative system, increasing error with 2nd-order Runge-Kutta. (c) Small energy error with variational integrator for dissipative system, large error with 1st-order Runge-Kutta. (d) No contact violation for box drop with rigid contact formulation, violation with MuJoCo's soft formulation.}
	\label{fig:physical_accuracy}
\end{figure}

The physical accuracy of the simulator is examined in four scenarios: constraint drift, energy conservation, energy dissipation, and contact violation. A comparison to commonly used alternative implementations is provided for reference. The results are displayed in Fig. \ref{fig:physical_accuracy}.

For the constraint drift in Fig. \ref{fig:physical_accuracy} (a), a three-link mechanism forming a kinematic loop is used with link lengths $l_1=1$m, $l_2=\scriptstyle{\frac{\sqrt{2}}{2}}$m, $l_3=1$m, masses $m_1=1$kg, $m_2=\scriptstyle{\frac{\sqrt{2}}{2}}$kg, $m_3=1$kg, and a step size $\Delta t=0.01$s. In minimal coordinates, the loop-closure constraints must be explicitly enforced. In non-variational integrators, the dynamics and constraints are formulated on an acceleration level. Without constraint stabilization \cite{baumgarte_stabilization_1972}, i.e., a spring-damper connection, the links of the mechanism start to drift apart. The explicit 4th-order Runge-Kutta-Munte-Kaas integrator \cite{hairer_geometric_2006} exhibits constraint drift without such stabilization. The 1st-order variational integrator prevents constraint drift entirely and does not require constraint stabilization.

The energy conservation in Fig. \ref{fig:physical_accuracy} (b) is evaluated on a frictionless double pendulum with link lengths $l=1$m, masses $m=1$kg, and a step size $\Delta t=0.01$s. \myupdate{Unlike variational integrators, strictly explicit or implicit Runge-Kutta integrators do not generally have tight bounds on the energy error for conservative mechanical systems, although certain methods, for example symmetric ones, do \cite{hairer_geometric_2006}.}
The explicit 2nd-order Runge-Kutta method (Heun's method) injects energy into the system, while the energy error stays bounded for the variational integrator.

Accurate energy dissipation is an important property, for example, in passivity-based control approaches \cite{music_passive_2018}. The dissipation behavior in Fig. \ref{fig:physical_accuracy} (c) is evaluated on a damped pendulum with link length $l=1$m, mass $m=1$kg, joint damping $d=\frac{1}{2}\text{N}\frac{\text{s}}{\text{m}}$ and a step size $\Delta t=0.1$s. Euler's method used for comparison shows poor dissipation behavior in drastically underdamping the pendulum, whereas the variational integrator demonstrates good dissipation performance after a small initial error.

Correct simulation of rigid contacts is crucial for transferring learned or optimized control policies from simulation to real systems. To compare rigid and soft contacts, a cube with edge length $l=0.5$m, mass $m=1$kg, and step size $\Delta t=0.01$s is dropped from a height (bottom-to-ground) $h=0.4$m. MuJoCo's default solver parameters ($\mathrm{solimp} = (0.9, 0.95, 0.001, 0.5, 2)$, $\mathrm{solref} = (0.02, 1)$) and default Euler integrator are used and result in a ground violation of $2.7$cm. We also analyzed drops from other heights up to 1m, which resulted in similar violations. In contrast, the rigid contact formulation with the variational integrator in maximal coordinates stops $43$$\mathrm{\mu}$m above the ground due to the interior-point formulation, which practically satisfies the constraint.

\subsection{Computational Complexity}\label{sec:eval_comp_compl}
\begin{figure}[!htb]
	\centering
	\input{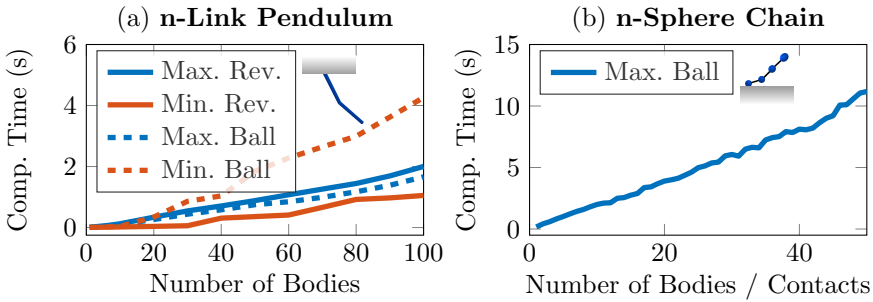}
	\caption{Computation time comparison for simulating 1000 time steps with our simulator (blue) compared to RigidBodyDynamics (red). (a) n-link pendulum with revolute (solid) and spherical (dashed) joints. (b) n-sphere chain with contacts and spherical joints.}
	\label{fig:linear_time}
\end{figure}

\begin{figure}[!htb]
	\centering
	\input{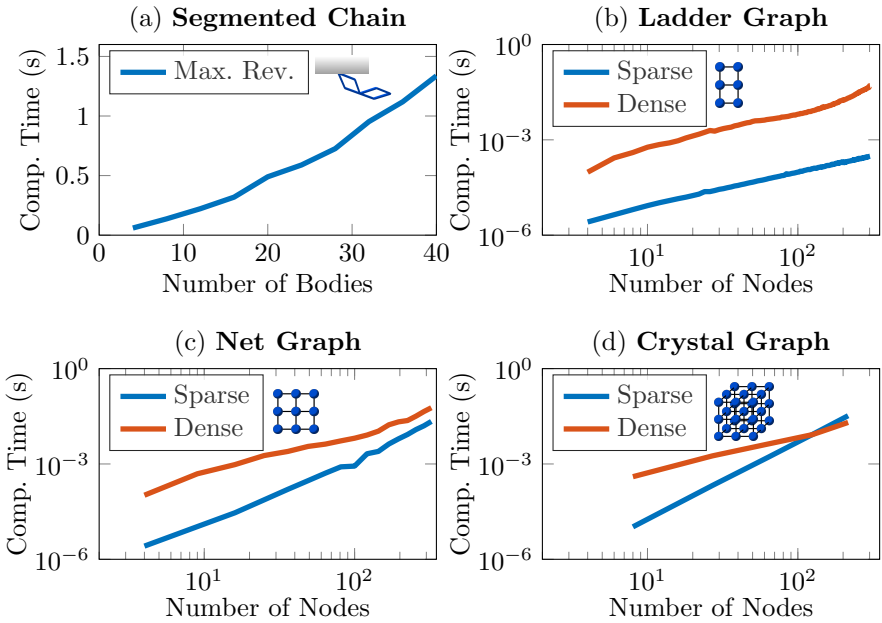}
	\caption{Evaluation and comparison of our methods (blue) for cyclic structures compared to standard dense implementations (red). (a) Simulation of 1000 time steps of a chain consisting of 4-link segments. (b) Solving of linear system from ladder graph. (c) Solving of linear system from net graph. (d) Solving of linear system from crystal graph.}
	\label{fig:loops}
\end{figure}

The evaluation of the computational complexity serves two purposes. We show that the complexity of the graph-based algorithms holds in practice, and by using these algorithms, maximal coordinates can achieve competitive timing results compared to minimal coordinates despite their larger dimension. For all timings, the best result of 100 samples is used to diminish right-skewing computer noise. The linear complexity of the simulator and a comparison to minimal coordinates are shown in Fig. \ref{fig:linear_time}. The performance for systems with kinematic loops and comparisons to a dense solver are displayed in Fig. \ref{fig:loops}.

Figure \ref{fig:linear_time} (a) compares the computation time of our simulator with the RigidBodyDynamics simulator for 1000 time steps of $n$-link pendulums with link length $l=1$m, mass $m=1$kg, and step size $\Delta t=0.01$s. The comparison is made for revolute and spherical (ball-and-socket) joints between the links. The main result is that despite the higher dimension of maximal coordinates, comparable computation times are achieved. Minimal coordinates are naturally faster for revolute joints, as they have the smallest number of degrees of freedom in minimal coordinates and the most constraints in maximal coordinates. On the other hand, spherical joints perform worse in minimal coordinates as these joints increase their state dimension and reduce the number of constraints in maximal coordinates. The linear complexity for both minimal and maximal coordinates becomes clearly visible.

In Fig. \ref{fig:linear_time} (b), the linear complexity for contacts is demonstrated for a chain of spheres with radius $r=0.25$m, mass $m=1$kg, and a step size $\Delta t=0.01$s. The spheres are connected by spherical joints. A comparison to RigidBodyDynamics is not possible due to the limited support of contacts. Besides the complexity, this experiment demonstrates the numerical robustness of the maximal-coordinate approach for treating contacts. We also implemented the experiment in MuJoCo. While faster, MuJoCo with default solver parameters consistently has contact violations of 10cm or more when simulating more than 20 spheres and fails to compute chains of more than 44 spheres. Therefore, a fair comparison is difficult.

The computation time for simulating 1000 time steps of a chain of 4-link segments is shown in Fig. \ref{fig:loops} (a). The links have a length $l=1$m, mass $m=1$kg, and the step size is $\Delta t=0.01$s. While no longer fully linear, the increase in computation time for this system is modest due to the reduction of fill-in. A simulation of this system with the RigidBodyDynamics simulator failed for more than one 4-link segment, even for smaller step sizes. Numerically, MuJoCo can simulate this system successfully, but with the default solver parameters, the explicit loop-closure constraints introduce high damping into the system, resulting in bad energy conservation behavior.

Comparisons of the sparse graph-based system solver with a dense one are shown in Fig. \ref{fig:loops} (c) - (d). The ladder graph consists of cycles with four nodes each, and two nodes are shared by two cycles, except for the first and last two nodes. The net graph is a square net of nodes, where inner nodes are part of four cycles, edge nodes part of two, and corner nodes part of one. The crystal graph is a three-dimensional cubic structure of nodes, where inner nodes are part of twelve cycles, surface nodes part of eight, edge nodes part of five, and corner nodes part of three. The graphs with $n$ nodes are represented by matrices with $6n$ entries. The sparse algorithms outperform the dense ones in most cases, often by more than two orders of magnitude. The crystal graph relates to a rather dense matrix, resulting in decreasing advantage of the spares approach over the dense one and slightly better performance for $6^3 = 216$ nodes. This last example is an extreme case. The algorithms are aimed at robotic systems, which typically have significantly fewer nodes and cycles, and the sparse performance is convincing for such systems.

\begin{figure}[!htb]
	\centering
	\input{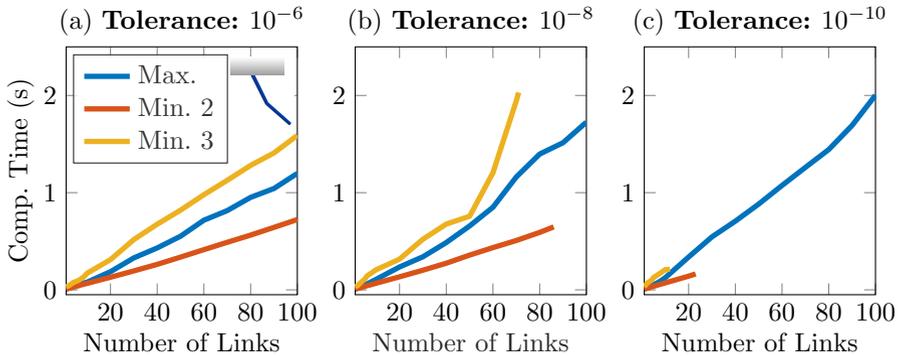}
	\caption{Performance comparison simulating 1000 time steps of n-link pendulums with varying solution tolerances. Comparison of our $1^{\text{st}}$-order algorithm (blue) to a $2^{\text{nd}}$-order (red) and a $3^{\text{rd}}$-order variational integrator (yellow). (a) Tolerance of $10^{-6}$. (b) Tolerance of $10^{-8}$. (c) Tolerance of $10^{-10}$.}
	\label{fig:varint_comp}
\end{figure}

\begin{figure}[!htb]
	\centering
	\input{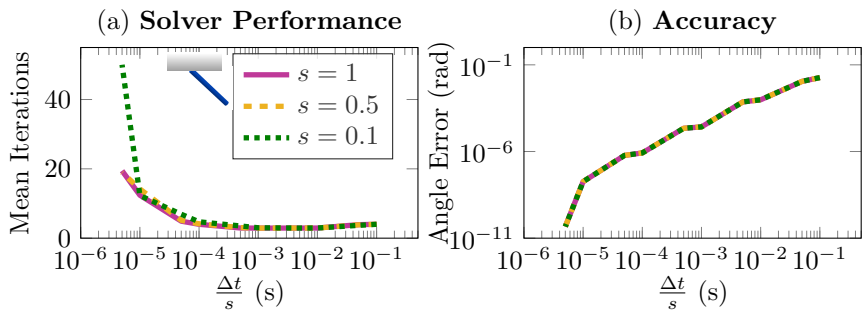}
	\caption{Evaluation of the solver performance for pendulums of scale $s=1$ (solid pink), $s=0.5$ (dashed yellow), and $s=0.1$ (dotted green) for scaled step sizes. (a) Mean iteration number. (b) Error of final angle.}
	\label{fig:condition_comp}
\end{figure}

\begin{figure}[!htb]
	\centering
	\input{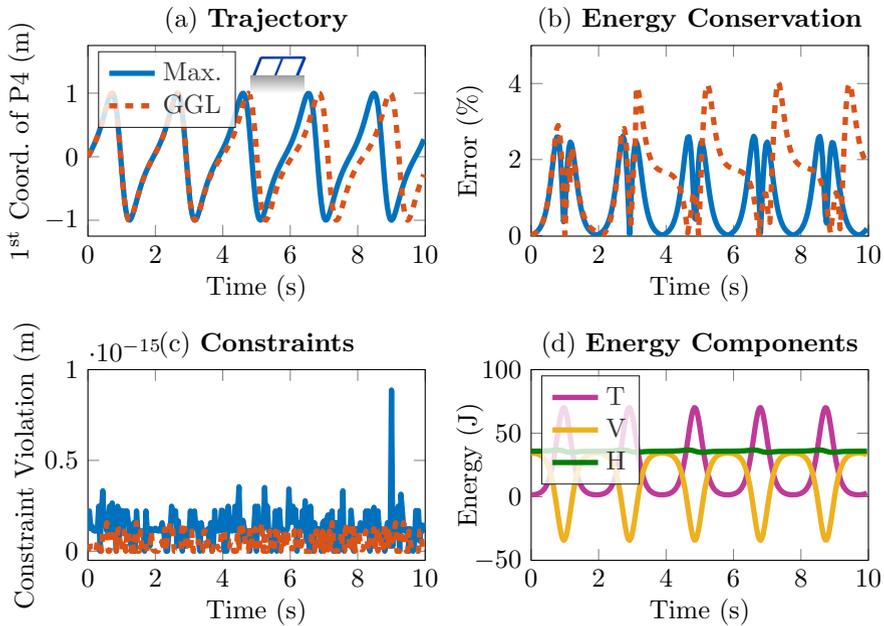}
	\caption{Comparison of the double-four-bar linkage benchmark system for our algorithm (blue) and the variational GGL method (red). (a) Simulation trajectory. (b) Energy error. (c) Constraint violation. (d) Energy components.}
	\label{fig:double4bar_comp}
\end{figure}

\myupdate{\subsection{Numerical Robustness}
We investigate three different scenarios regarding the numerical robustness of the simulator. A comparison with two state-of-the-art variational integrators in minimal coordinates is made to demonstrate the ability to simulate systems with decreasing solution tolerance. The results are displayed in Fig. \ref{fig:varint_comp}. Since constrained mechanical systems are known to become ill-conditioned for decreasing step sizes \cite{petzold_numerical_1986,cardenal_multi_1999}, we show the ability to simulate systems for reasonable step sizes in Fig. \ref{fig:condition_comp}. A double-four-bar linkage---a commonly-used benchmark problem that exhibits singular constraint configurations---is simulated, and the results are compared to another variational integrator for constrained systems in Fig. \ref{fig:double4bar_comp}.}

\myupdate{In Fig. \ref{fig:varint_comp}, we compare our algorithm to state-of-the-art second- and third-order variational integrators with the same n-link pendulum we use for the timing comparison in Sec. \ref{sec:eval_comp_compl}. Only revolute joints and loop-free structures are tested as ball-and-socket joints, and loop-closures were, to the best of our knowledge, not part of the implementations of these integrators. To demonstrate the robustness of our algorithm, we run simulations with solution tolerances of $10^{-6}$, $10^{-8}$, and $10^{-10}$ for the Newton methods in all algorithms. All three algorithms display the theoretical linear computational complexity, and despite the higher dimensionality of the maximal-coordinate algorithm, we achieve comparable timing results. However, while our algorithm successfully simulates the n-link pendulums for all tolerances, the minimal-coordinate integrators fail for smaller tolerances and an increasing number of links, hinting at potential numerical issues in minimal coordinates.}

\myupdate{Generally, large simulation step sizes are desirable for fast simulations. However, smaller step sizes are required for physically smaller systems to capture their high-frequency motions. Moreover, smaller step sizes may be necessary to achieve highly accurate results. By simulating a physically scaled pendulum with different step sizes, we determine the time-step range for which we can successfully simulate the systems. The pendulum of scale $s\in[1,0.5,0.1]$ has a mass $m=s$, length $l=s^2$, joint inertia $J=\frac{1}{3}ml^2$, and gravitational acceleration $g=9.81$. Accordingly, the period of the pendulum is $T=2\pi\sqrt{\frac{2}{3}\frac{l}{g}}=2\pi\sqrt{\frac{2}{3}\frac{1}{g}}s$, i.e., proportional to the scale $s$. The pendulum is initialized at an angle and angular velocity $[\theta_0,\dot{\theta}_0]=[\frac{\pi}{2},0]$ and the simulation time is $\frac{T}{2}$, i.e., a half swing. The results in Fig. \ref{fig:condition_comp} are plotted for a step size scaled by $s$ to account for the different frequencies of the systems. An increase in the average iteration number per time step for smaller step sizes can be seen in Fig. \ref{fig:condition_comp} (a), which is consistent with the increased ill-conditioning of the systems. Nonetheless, the simulation is successful and achieves accurate results even for small step sizes. The accuracy is measured as $\mathrm{norm}(-\frac{\pi}{2}-\theta_T)$, i.e., the error of the angle at the last time step.}

\myupdate{We simulate the double-four-bar linkage, a common benchmark system \cite{gonzalez_benchmarking_2006}, to evaluate the performance of our simulator since the system encounters configuration singularities during simulation when the bars are in parallel. We compare our algorithm to the GGL method in \cite{kinon_ggl_2023} and use the same notation and parameters for the system. The simulation step size is $\Delta t=0.01$s. As Fig. \ref{fig:double4bar_comp} (a) and (b) show, our method simulates the system with a bounded error on the energy, whereas the GGL method incurs an increasing energy error, which also leads to an increasingly wrong trajectory. Regarding the constraint satisfaction shown in Fig. \ref{fig:double4bar_comp} (c), both algorithms achieve very low constraint violations, although GGL performs better, potentially because the GGL method enforces the constraints also on a velocity level. Figure \ref{fig:double4bar_comp} (d) shows the kinetic, potential, and total energy of the system as a reference for other benchmarks. We also ran the benchmark for a step size $\Delta t=0.001$s, where almost no difference exists between the two methods. A computation time comparison between uncompiled Matlab and compiled Julia code is difficult, and the GGL method is not designed for numerical efficiency, but it is still worth mentioning that our algorithm was more than 100 times faster than the GGL method.}

\subsection{Application Examples}	
\begin{figure}[!htb]
	\centering
	\input{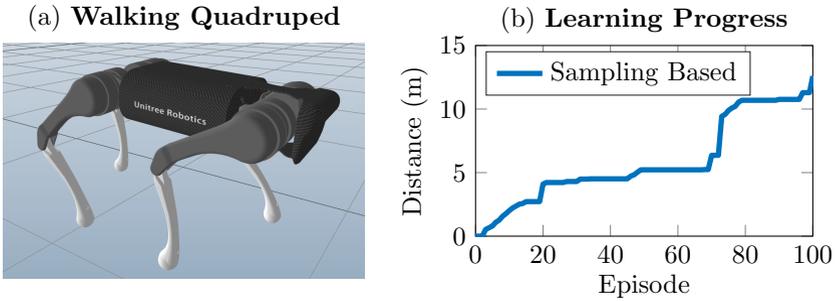}
	\caption{(a) The trained walking quadruped. (b) Progress for learning to walk.}
	\label{fig:walker}
\end{figure}

\begin{figure}[!htb]
		\centering
		\input{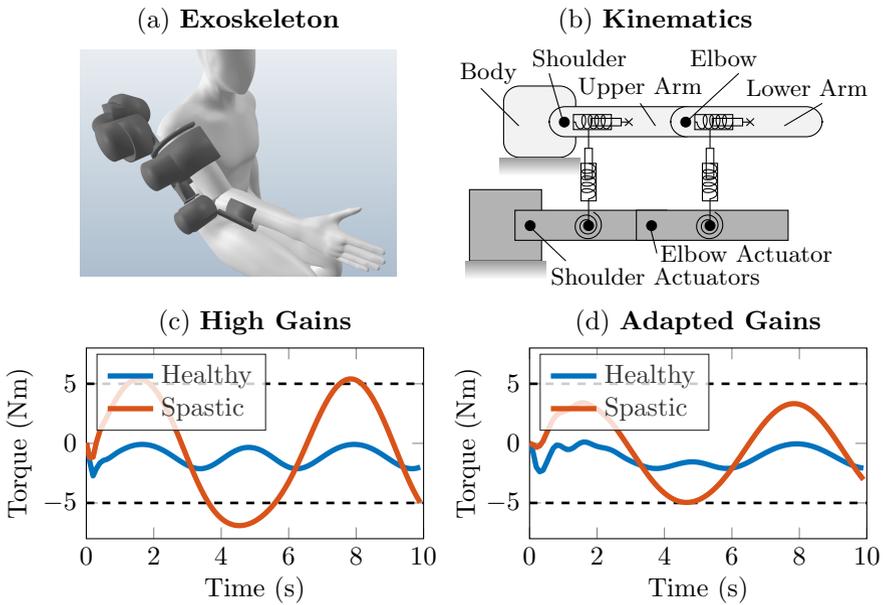}
		\caption{(a) The donned exoskeleton in simulation. (b) Schematics of the exoskeleton kinematics (not to scale) showing the joints and kinematic loops. Offset between exoskeleton and body exaggerated. (c) Elbow torques during rehabilitation task for high gains. Within limits for healthy patient, exceeding limits for spastic patient. (d) Elbow torques for adapted gains within the limits for both patients.}
		\label{fig:exo}
	\end{figure}

This article focuses on the theoretical foundation for building efficient and physically accurate maximal-coordinate simulators. However, an implementation with a basic interior-point method can already be used for real-world applications. As such application examples, we show that the control parameters for a quadrupedal robot can be learned with a simple sampling-based approach and how controller gains of an exoskeleton can be personalized to impaired patients in simulation to avoid injury. The quadrupedal robot and learning progress are displayed in Fig. \ref{fig:walker}. The exoskeleton and resulting torques for original and adapted controller gains are visualized in Fig. \ref{fig:exo}.

We use the Unitree A1 quadrupedal robot for sampling-based learning. The simulation step size is $\Delta t = 0.001$s. The learning algorithm is stated in Appendix \ref{sec:app_applications}. For each episode, a random set of control parameters in the vicinity of the current set of parameters is chosen, and a simulation rollout is performed for 10 seconds. In case of progress, i.e., the walking distance increased, this set of parameters is selected as the starting point for new sample draws. The robot learns to walk a distance of more than 12m (average velocity of $1.2\frac{\text{m}}{\text{s}}$) in 100 episodes. Due to the rigid contacts of the simulation, a successful transfer of the learned control parameters to a real system is more likely than with an incorrectly soft contact model. 

For the gain tuning, a 4-degrees-of-freedom (DoF) exoskeleton for an arm is used. Three DoF actuate the shoulder and upper arm, and one DoF the elbow and lower arm. The attachments of the upper and lower arm are modeled as two 3-DoF spring-damper joints with two translational and one rotational DoF. As a result, there are two connected kinematic loops in the mechanism. Commonly, attachments between the exoskeleton and the human body are modeled as pure spring-damper connections without joints to avoid such kinematic loops, for example, in \cite{kuhn_dynamics_2018}. However, such models are not necessarily correct. For the exoskeleton application, we assume a healthy and a spastic patient that perform a rehabilitation routine in the exoskeleton. The shoulder and elbow flexion/extension joints are supposed to follow sinusoidal trajectories (see Appendix \ref{sec:app_applications}). A limit of $5$Nm is assumed to be a comfortable elbow torque for the patients. While the original gains stay within these limits for the healthy patient, they are exceeded for the spastic patient. By tuning the gains in simulation, an adapted set of gains adhering to the torque limits for both patients is found. With this approach, uncomfortable and potentially harmful tuning on a real patient can be avoided or at least reduced. The modeling accuracy, including proper kinematic loops, should provide closer estimates of the controller gains for the real system.

\section{Conclusions}\label{sec:conclusions}
This article introduces a maximal-coordinate variational integrator and efficient graph-based solver for simulating mechanical systems with common components such as springs and dampers, actuated joints, and contacts with friction. Besides the theoretical formulation of the integrator and solver algorithms, an application-ready implementation of the simulator is provided as open-source code.

Building maximal-coordinate simulators on variational integrators is useful not just for the conservation properties and physical accuracy but also for avoiding constraint drift in the naturally constrained formulation. The increased state dimension is treated with efficient numerical solver algorithms, which reduce the computational complexity in theory and render the simulator useable in practice for various applications. Additionally, it appears that the formulation in maximal coordinates increases the numerical robustness and allows for the simulation of systems with contacts or kinematic loops that other simulators in minimal coordinates fail to compute. 

Because of the simple modular formulation in maximal coordinates, additional components can be added to the integrator, such as a nonlinear friction model, joint limits, or physically correct elastic contacts. The graph-based nature of the solver algorithms also opens up the possibility of parallelizing computations on different branches of a graph.

\backmatter

\bmhead{Acknowledgments}
The authors would like to thank Petar Bevanda for his help in preparing the manuscript, as well as Marko Galic and Jana Janeva for their implementation help.

\section*{Declarations}

\begin{itemize}
\item Competing interests: The authors have no competing interests to declare.
\end{itemize}

\begin{appendices}

\section{Quaternions}\label{sec:app_quaternions}
We use quaternions as rotation representations since they are globally non-singular, as opposed to three-parameter representations, and numerically efficient, having only four parameters, unlike, for example, rotation matrices with nine parameters.

\subsection{Notation}

We write quaternions as a stacked vector,
\begin{equation}\label{eqn:quat}
\bs{q}
=
\begin{bmatrix}
q_s \\ q_{v_1} \\ q_{v_2} \\ q_{v_3}
\end{bmatrix}
= 
\begin{bmatrix}
q_s \\ \bs{q}_v
\end{bmatrix}\in\myupdate{\mathbb{H}},
\end{equation}
where $q_s$ and $\bs{q}_v$ are the scalar and vector parts, respectively. We follow the Hamilton convention with a local-to-global rotation action. In this convention, a quaternion $\bs{q}$ maps vectors from the local to the global frame, whereas its inverse maps from the global to the local frame.

Notation \eqref{eqn:quat} allows for a simple formulation of the basic operations conjugate, inverse, and multiplication:
\begin{subequations}
	\begin{alignat}{2}
	&\text{Conjugate: } ~~&&~~\bs{q}\hC = 
	\begin{bmatrix} 
	q_s \\ 
	-\qno{v} 
	\end{bmatrix} \\
	&\text{Inverse: } ~~&&~\bs{q}^{-1} = \frac{\bs{q}\hC}{\lVert\bs{q}\rVert}\\
	&\text{Multiplication: } ~~&&\bs{q} \cdot \bs{p} = 
	\begin{bmatrix}
	q_{s} p_{s} - \qno{v}\hT \bs{p}_v 
	\\q_{s}\bs{p}_v + p_{s}\qno{v} + \qno{v} \times \bs{p}_v
	\end{bmatrix}
	\end{alignat}
\end{subequations}
Note that, for unit quaternions, $\bs{q}\hC = \bs{q}^{-1}$, and that the $\times$ operator indicates the standard cross product of two vectors. 

Three other common operations are expanding a vector $\bs{x} \in \mathbb{R}^3$ into a quaternion, retrieving the vector part from a quaternion, and constructing a skew-symmetric matrix from a vector $\bs{x} \in \mathbb{R}^3$ to form the cross product as a matrix-vector product:
\begin{subequations}
	\begin{alignat}{2}
	&\text{Expand vector: } ~~&&\bs{x}^{\wedge} = 
	\begin{bmatrix} 
	0 \\ 
	\bs{x} 
	\end{bmatrix}\\
	&\text{Retrieve vector: } ~~&&\bs{q}^{\vee} = \qno{v}
	\\
	&\text{Skew-symmetric matrix: } ~~&&\bs{x}^{\times} = \begin{bmatrix}0 & -x_3 & x_2\\ x_3 & 0 & -x_1\\ -x_2 & x_1 & 0\end{bmatrix} 
	\end{alignat}
\end{subequations}
The cross product of two vectors can then be written as $\xno{1}\times\xno{2} = \xno{1}^{\times}\xno{2}$.

To simplify calculations with quaternions, we introduce the following  four matrices with the identity matrix $\bs{I}_{3\times 3} \in \mathbb{R}^{3\times3}$:
\begin{subequations}
	\begin{alignat}{2}
	\bs{T} &= \begin{bmatrix}1 &~~ \bs{0}\hT\\\bs{0} &~~ -\bs{I}_{3\times 3}\end{bmatrix} &&~~ \in \mathbb{R}^{4\times4},\\
	\Lmat{\bs{q}} &= \begin{bmatrix}q_s &~~ -\qno{v}\hT\\\qno{v} &~~ q_s \bs{I}_{3\times 3} + \qno{v}^{\times} \end{bmatrix} &&~~ \in \mathbb{R}^{4\times4},\\
	\Rmat{\bs{q}} &= \begin{bmatrix}q_s &~~ -\qno{v}\hT\\\qno{v} &~~ q_s \bs{I}_{3\times 3} - \qno{v}^{\times} \end{bmatrix} &&~~ \in \mathbb{R}^{4\times4},\\
	\bs{V} &= \begin{bmatrix}\bs{0} &~~ \bs{I}_{3\times 3}\end{bmatrix} &&~~ \in \mathbb{R}^{3\times4}.
	\end{alignat}
\end{subequations}
With these matrices, all required quaternion operations can be written as matrix-vector products for which the standard rules of linear algebra hold:
\begin{subequations}
	\begin{align}
	\bs{q}_1 \cdot \bs{q}_2 &= \Lmat{\bs{q}_1}\bs{q}_2 = \Rmat{\bs{q}_2}\bs{q}_1,\\
	\bs{q}^{-1} &= \bs{T}\bs{q},\\
	\bs{q}^{\vee} &= \bs{V}\bs{q},\\
	\bs{x}^{\wedge} &= \bs{V}\hT\bs{x}.
	\end{align}
\end{subequations}

The rotation of a vector $\bs{x}$ can be expressed as
\begin{equation}
\left(\bs{q}\cdot\bs{x}^{\wedge}\cdot\bs{q}^{-1}\right)^{\vee} = \bs{Q}(\bs{q})\bs{x},
\end{equation}
where $\bs{Q}(\bs{q})=\bs{V} \RTmat{\bs{q}} \Lmat{\bs{q}} \bs{V}\hT$ is the rotation matrix formed from $\bs{q}$.

\subsection{Derivatives}
While quaternions have four parameters, rotations only have three degrees of freedom, and so we use specialized derivatives for quaternion functions following \cite{brudigam_linear-time_2020,jackson_planning_2021}.

The rotational gradient of a quaternion-dependent scalar function $f(\bs{q})$ is defined as
\begin{equation}
	\nabla_{\bs{q}}^\mathrm{r}f(\bs{q}) = \bs{V}\Lmat{\bs{q}}\hT\nabla_{\bs{q}}f(\bs{q}),
\end{equation}
and the rotational Jacobian of a vector-valued function $\bs{f}(\bs{q})$ as
\begin{equation}
	\frac{\partial \bs{f}(\bs{q})}{\partial^\mathrm{r}\bs{q}} = \frac{\partial \bs{f}(\bs{q})}{\partial\bs{q}}\Lmat{\bs{q}}\bs{V}\hT.
\end{equation}

\subsection{Properties for Dynamics Descriptions}
Using quaternions in the description of dynamical systems requires some attention.
\paragraph*{Angular Velocity}
In continuous time, the quaternion angular velocity is defined as
\begin{equation}\label{eqn:quatangvel}
	\bar{\bs{\omega}} = \begin{bmatrix}
		\bar{\omega}_s\\
		\bs{\omega}
	\end{bmatrix} = 2\LTmat{\bs{q}}\dot{\bs{q}},
\end{equation}
where $\bar{\omega}_s = 0$. But when using a first-order approximation of $\dot{\bs{q}}$,
\begin{equation}\label{eqn:quatapprox}
	\dot{\bs{q}}_k = \frac{\qno{k+1}-\qno{k}}{\dt},
\end{equation}
\myupdate{we obtain a discretized quaternion angular velocity $\bar{\bs{\omega}}_k$, generally with a scalar part} $\bar{\omega}_{k,s}\neq 0$. Therefore, \myupdate{$\bar{\bs{\omega}}_k$} is defined so that given $\bs{q}_k$ and $\bs{\omega}_k$, $\bs{q}_{k+1}$ maintains unit norm. 

\myupdate{Given} the discretized angular velocity
\begin{align}
	\bs{\omega}_k &= \left(2\myupdate{\Lmat{\qno{k}}\hT}\frac{\qno{k+1}-\qno{k}}{\dt}\right)^{\vee}\nonumber\\
	&=\frac{2}{\dt}\left(\myupdate{\Lmat{\qno{k}}\hT}\qno{k+1}\right)^{\vee},
\end{align}
\myupdate{we define the discretized quaternion angular velocity as 
\begin{equation}
	\bar{\bs{\omega}}_k = \frac{2}{\dt}\Lmat{\qno{k}}\hT\qno{k+1}.
\end{equation}}
Since $\myupdate{\Lmat{\qno{k}}\hT}\qno{k+1}$ is \myupdate{an orientation and} must have unit norm, the constraint on $\bar{\bs{\omega}}_k$ is
\begin{equation}
	\left\lVert \frac{\dt}{2}\bar{\bs{\omega}}_k\right\rVert^2 = \left(\frac{\dt}{2}\right)^2\bar{\omega}_{k,s}^2 + \left(\frac{\dt}{2}\right)^2\bs{\omega}_k\hT\bs{\omega}_k = 1.
\end{equation}
As a result,
\begin{equation}
	\bar{\bs{\omega}}_k=\begin{bmatrix}\bar{\omega}_{k,s}\\ \wno{k}\end{bmatrix}=\begin{bmatrix}\sqrt{\left(\tfrac{2}{\dt}\right)^2 - \wno{k}\hT\wno{k}}~\\ \wno{k}\end{bmatrix}.
\end{equation}
\myupdate{Note that $\bar{\omega}_{k,s} = \frac{2}{\dt}$ for $\bs{\omega}=0$. This difference to the continuous-time definition simplifies the integrator derivation and implementation.}

\paragraph*{Virtual Work}
\myupdate{In the Lagrange-d'Alembert principle, the virtual work for external forces and torques is
\begin{equation}
	\delta W  = \begin{bmatrix} \bs{\mathrm{f}}(\bs{z}) & \bs{\tau}(\bs{z}) \end{bmatrix}\hT \delta \bs{z},
\end{equation}
where $\delta \bs{z}$ represents variations of the trajectory. In our case, the force $\bs{\mathrm{f}}(\bs{z})=\bs{\mathrm{f}}$ is not state-dependent, and the torque $\bs{\tau}(\bs{z}) = 2\Lmat{\bs{q}} \bs{V}\hT \bs{\tau}$ depends on a body's current orientation. In the integrator derivation, gradients represent variations $\delta \bs{z}$.}
\newpage
\section{Kinematic Joints}\label{sec:app_joints}
\begin{figure}[!htb]
	\centering
	\begin{tikzpicture}
	\node[inner sep=0pt] (ref) at (-1,0){ };

	\coordinate (w) at (1,0);
	\node at ($(w)+(-0.2,2.3)$) {(a)};
	\draw[->,draw=blue!80,thick] (w) -- ($(w)+(0.0,0.42)$);
	\draw[->,draw=blue!80,thick] (w) -- ($(w)+(0.42,0.0)$);
	\draw[->,draw=blue!80,thick] (w) -- ($(w)+(-0.15,-0.3)$);

	\draw[->,draw=black,thick] (w) -- ($(w)+(1,2)$);
	\node at ($(w)+(0.2,1.1)$) {$\xtx{a}$};
	\draw[->,draw=black,thick] ($(w)+(1,2)$) -- ($(w)+(1.5,2)$);
	\node at ($(w)+(1.2,2.2)$) {$\btx{p}{a}$};
	\draw[->,draw=black,thick] (w) -- ($(w)+(2,1)$);
	\node at ($(w)+(1.2,0.3)$) {$\xtx{b}$};
	\draw[->,draw=black,thick] ($(w)+(2,1)$) -- ($(w)+(2,1.5)$);
	\node at ($(w)+(2.3,1.2)$) {$\btx{p}{b}$};
	\draw[->,draw=black,thick] ($(w)+(1.5,2)$) -- ($(w)+(2,1.5)$);
	\node at ($(w)+(2.0,1.9)$) {$\btx{g}{T}$};

	\coordinate (w) at (5.5,0);
	\node at ($(w)+(-0.2,2.3)$) {(b)};
	\draw[->,draw=blue!80,thick] (w) -- ($(w)+(0.0,0.42)$);
	\draw[->,draw=blue!80,thick] (w) -- ($(w)+(0.42,0.0)$);
	\draw[->,draw=blue!80,thick] (w) -- ($(w)+(-0.15,-0.3)$);

	\coordinate (l1) at ($(w)+(0.5,1.5)$);
	\draw[->,draw=black,thick] (l1) -- ($(l1)+(-0.16,0.4)$);
	\draw[->,draw=black,thick] (l1) -- ($(l1)+(0.4,0.16)$);
	\draw[->,draw=black,thick] (l1) -- ($(l1)+(-0.15,-0.3)$);
	\node at ($(w)+(-0.05,0.9)$) {$\qtx{a}$};

	\draw ($(w)+(-0.07,0.07)$) to[out=150,in=-80 ] ($(l1)+(0.05,-0.05)$);

	\coordinate (l2) at ($(w)+(3,0.5)$);
	\draw[->,draw=black,thick] (l2) -- ($(l2)+(0.23,0.36)$);
	\draw[->,draw=black,thick] (l2) -- ($(l2)+(0.36,-0.23)$);
	\draw[->,draw=black,thick] (l2) -- ($(l2)+(-0.15,-0.3)$);
	\node at ($(w)+(1,0.15)$) {$\qtx{b}$};

	\draw ($(w)+(0.07,0.07)$) to[out=50,in=-50 ] ($(l2)+(0.03,-0.09)$);

	\coordinate (loff) at ($(w)+(2,1.2)$);
	\draw[->,draw=black,thick] (loff) -- ($(loff)+(0.0,0.42)$);
	\draw[->,draw=black,thick] (loff) -- ($(loff)+(0.42,0.0)$);
	\draw[->,draw=black,thick] (loff) -- ($(loff)+(-0.15,-0.3)$);

	\draw ($(l1)+(0.07,0.09)$) to[out=60,in=160 ] ($(loff)+(-0.07,-0.02)$);
	\node at ($(w)+(1.5,1.8)$) {$\qtx{off}{}$};
	\draw ($(loff)+(0.07,-0.07)$) to[out=-60,in=140 ] ($(l2)+(-0.07,-0.02)$);
	\node at ($(w)+(2.7,0.9)$) {$\btx{g}{R}$};

\end{tikzpicture}
	\caption{Visualization of the two general constraints. (a) The relationship of the translational constraint components. (b) The relationship of the rotational constraint components.}
	\label{fig:constraints}
\end{figure}
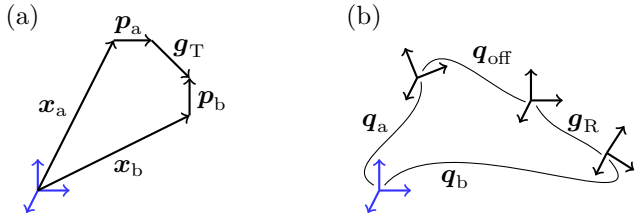

Most of the common joints can be defined by composing a general translational constraint function and a general rotational constraint function. A visualization of the two constraints is given in Fig. \ref{fig:constraints}. This representation based on two general constraint functions also allows for easy extraction of minimal coordinates.

\subsection{Translational Constraint}
The general translational constraint function describes the relative distance of two points relative to the origins of the local frames of two bodies:
\myupdate{\begin{equation}
	\bs{g}\mrml{T} = \bs{Q}(\bs{q}\mrml{a})\hT\left(\left(\bs{x}\mrml{b} + \bs{Q}(\bs{q}\mrml{b})\bs{p}\mrml{b}\right) - \left(\bs{x}\mrml{a} + \bs{Q}(\bs{q}\mrml{a})\bs{p}\mrml{a}\right)\right).
\end{equation}} 
The vectors $\bs{p}\mrml{a}$ and $\bs{p}\mrml{b}$\myupdate{, pointing from the center of mass to the joint,} are defined in the respective local frames, and the resulting relative distance is defined in body $\mathrm{a}$'s frame.

In case of only a single body directly connected to the global frame, i.e., $\bs{x}\mrml{a}=\bs{0}$ and $\bs{q}\mrml{a}=\mathbbm{1}$ (identity quaternion), we obtain
\begin{equation}
	\bs{g}\mrml{T} = \bs{x}\mrml{b} + \myupdate{\bs{Q}(\bs{q}\mrml{b})\bs{p}\mrml{b}} - \bs{p}\mrml{a}.
\end{equation} 

\subsection{Rotational Constraint}
The general rotational constraint function describes the relative distance of two local frames of two bodies, including a possible offset:
\begin{equation}
	\bs{g}\mrml{R} = \myupdate{\bs{V}\Lmat{\bs{q}\mrml{off}}\hT\Lmat{\bs{q}\mrml{a}}\hT\bs{q}\mrml{b}}.
\end{equation}
The offset quaternion $\bs{q}\mrml{off}$ is defined in body $\mathrm{a}$'s frame, and the resulting relative rotation is defined in body $\mathrm{a}$'s frame as well.

In case of only a single body directly connected to the global frame, i.e., $\bs{q}\mrml{a}=\mathbbm{1}$, we obtain
\begin{equation}
	\bs{g}\mrml{R} = \myupdate{\bs{V}\Lmat{\bs{q}\mrml{off}}\hT\bs{q}\mrml{b}}.
\end{equation} 

\subsection{Composite Constraints and Minimal Coordinates}
To obtain actual joint constraints, the general constraint functions are multiplied with a selection matrix $\bs{D}$ indicating the desired constraints. Multiplying the constraint functions with the nullspace matrix $\bs{C}$ of $\bs{D}$ yields the corresponding minimal coordinates. 

The selection matrix $\bs{D}$ is calculated by performing singular value decomposition on a skew-symmetric matrix formed from a vector $\bs{V}_3$:
\begin{equation}
	\mathrm{svd}(\bs{V}_3^{\times}) = \bs{U} \bs{\Sigma} \bs{V}\mrmu{T}.
\end{equation}
The matrix $V$ contains both the original vector $\bs{V}_3$ (sign-adjustment might be necessary) and two perpendicular vectors $\bs{V}_1$ and $\bs{V}_2$:
\begin{equation}
	\bs{V} = \begin{bmatrix}
		\bs{V}_1 ~& \bs{V}_2 ~& \bs{V}_3
	\end{bmatrix} = \begin{bmatrix}
		\bs{V}_{1:2} ~& \bs{V}_3
	\end{bmatrix}
\end{equation}
Using the matrices $\bs{V}_{1:2}$ and $\bs{V}_3$, a number of constraint functions $\bs{D}\mrmu{T}\bs{g}$ can be created, with the meaning depending on $\bs{D}$. The different options for $\bs{D}$ and the corresponding nullspace matrices $\bs{C}$ are stated in Table \ref{tab:constraint_selection}.
\begin{table}[htp!]
	\centering
	\begin{tabular}{| l | l | l |}
		\hline
		Selection Matrix $\bs{D}$ & Nullspace Matrix $\bs{C}$ &  Description \\
		\hline\hline
		$\bs{0}_{3\times 0} \in \mathbb{R}^{3\times0}$ & $\bs{V}$  or $\bs{I}_{3\times 3}\in \mathbb{R}^{3\times3}$ & No restrictions \\
		$\bs{V}_3 \in \mathbb{R}^{3\times1}$ & $\bs{V}_{1:2} \in \mathbb{R}^{3\times2}$ & Restricted to plane $\bs{V}_{1:2}$ \\
		$\bs{V}_{1:2} \in \mathbb{R}^{3\times2}$ & $\bs{V}_3 \in \mathbb{R}^{3\times1}$ & Restricted to axis $\bs{V}_3$ \\
		$\bs{V}$  or $\bs{I}_{3\times 3}\in \mathbb{R}^{3\times3}$ & $\bs{0}_{3\times 0} \in \mathbb{R}^{3\times0}$ &  Fully restricted\\
		\hline
	\end{tabular}
	\caption{Constraint selection and nullspace matrices.}
	\label{tab:constraint_selection}
\end{table}

Mechanical joints are created by using different selection and nullspace matrices for the translational and rotational constraints and stacking everything as
\begin{subequations}
	\begin{align}
		\bs{D}\mrmu{T}\bs{g} &= \begin{bmatrix}
			\bs{D}\mrml{T} & 0 \\ 0 & \bs{D}\mrml{R}
		\end{bmatrix}\mrmu{T}\begin{bmatrix}
			\bs{g}\mrml{T}\\\bs{g}\mrml{R}
		\end{bmatrix}=\begin{bmatrix}
			\bs{D}\mrml{T}\bs{g}\mrml{T}\\\bs{D}\mrml{R}\bs{g}\mrml{R}
		\end{bmatrix},\\
		\bs{C}\mrmu{T}\bs{g} &= \begin{bmatrix}
			\bs{C}\mrml{T} & 0 \\ 0 & \bs{C}\mrml{R}
		\end{bmatrix}\mrmu{T}\begin{bmatrix}
			\bs{g}\mrml{T}\\\bs{g}\mrml{R}
		\end{bmatrix}=\begin{bmatrix}
			\bs{C}\mrml{T}\bs{g}\mrml{T}\\\bs{C}\mrml{R}\bs{g}\mrml{R}
		\end{bmatrix}.
	\end{align}
\end{subequations}

A list of mechanical joints that can be created with this composition is given in Table \ref{tab:joint_list}.
\begin{table}[htp!]
	\centering 
	\begin{tabular}{|l| l| l| l| l|}
		\hline
		Joint Name 							& $\bs{D}\mrml{T}$	& $\bs{D}\mrml{R}$ & $\bs{C}\mrml{T}$	& $\bs{C}\mrml{R}$ \\
		\hline\hline
		Fixed								& $\bs{I}_{3\times 3}$		& $\bs{I}_{3\times 3}$ & $\bs{0}_{3\times 0}$		& $\bs{0}_{3\times 0}$ \\
		Prismatic							& $\bs{V}_{1:2}$		& $\bs{I}_{3\times 3}$ & $\bs{V}_3$		& $\bs{0}_{3\times 0}$ \\
		Planar, fixed orientation			& $\bs{V}_3$		& $\bs{I}_{3\times 3}$ & $\bs{V}_{1:2}$		& $\bs{0}_{3\times 0}$ \\
		Fixed orientation					& $\bs{0}_{3\times 0}$		& $\bs{I}_{3\times 3}$ & $\bs{I}_{3\times 3}$		& $\bs{0}_{3\times 0}$ \\
		Revolute (Hinge)					& $\bs{I}_{3\times 3}$		& $\bs{V}_{1:2}$ & $\bs{0}_{3\times 0}$		& $\bs{V}_3$ \\
		Cylindrical							& $\bs{V}_{1:2}$		& $\bs{V}_{1:2}$ & $\bs{V}_3$		& $\bs{V}_3$ \\
		Planar, rotation along axis			& $\bs{V}_3$		& $\bs{V}_{1:2}$ & $\bs{V}_{1:2}$		& $\bs{V}_3$ \\
		Rotation along axis, free movement	& $\bs{0}_{3\times 0}$		& $\bs{V}_{1:2}$ & $\bs{I}_{3\times 3}$		& $\bs{V}_3$ \\
		Spherical (ball-and-socket)			& $\bs{I}_{3\times 3}$		& $\bs{0}_{3\times 0}$ & $\bs{0}_{3\times 0}$		& $\bs{I}_{3\times 3}$ \\
		Cylindrical, free orientation		& $\bs{V}_{1:2}$		& $\bs{0}_{3\times 0}$ & $\bs{V}_3$		& $\bs{I}_{3\times 3}$ \\
		Planar, free orientation			& $\bs{V}_3$		& $\bs{0}_{3\times 0}$ & $\bs{V}_{1:2}$		& $\bs{I}_{3\times 3}$ \\
		Floating (unconstrained)			& $\bs{0}_{3\times 0}$		& $\bs{0}_{3\times 0}$ & $\bs{I}_{3\times 3}$		& $\bs{I}_{3\times 3}$ \\
		\hline
	\end{tabular}
	\caption{List of joints made of the two general constraint functions.}
	\label{tab:joint_list}
\end{table}

Joints with $\bs{D}\mrml{R} = \bs{V}_3$ do not appear to have any physical meaning. There are also joints that cannot be described by this composition, for example, helical joints (screws). Nonetheless, they can still be formulated as an equality constraint.
\newpage

\myupdate{\section{Friction Dynamics}\label{sec:app_friction}
The friction forces in a mechanism are derived from the maximum dissipation principle \cite{preclik_maximum_2018} with a  linearized friction cone \cite{stewart_implicit_1996}. The principle states that the energy dissipation rate of the bodies in contact is maximized through friction.}

\myupdate{Since friction acts on moving bodies, the energy dissipation rate is the time derivative of the kinetic energy. As the kinetic energy for all bodies of a mechanism is the sum of the kinetic energies of all individual bodies,
\begin{equation}
	\mathcal{T}(\dot{\bs{z}}) = \sum_{n=1}^{n_\mathrm{b}}\mathcal{T}(\dot{\bs{z}}_{\mathrm{n}}),
\end{equation}
the energy dissipation is the sum of the dissipation of all individual bodies. The energy dissipation for the contact point on a single body is
\begin{subequations}\label{eqn:fric_deriv}
	\begin{align}
		\frac{\mathrm{d}}{\mathrm{d}t}\mathcal{T}(\dot{\bs{z}}) &= \frac{\mathrm{d}}{\mathrm{d}t}\left(\frac{1}{2}\bs{v}\hT \bs{M}\bs{v} + \frac{1}{2}\bs{\omega}\hT \bs{J}\bs{\omega}\right)\\
		&=\bs{v}\hT \bs{M}\dot{\bs{v}} + \bs{\omega}\hT \bs{J}\dot{\bs{\omega}}\\
		&=\bs{v}\hT \bs{\mathrm{f}} + \bs{\omega}\hT \bs{\tau}\\
		&=\bs{v}\hT \bs{\mathrm{f}} + \bs{\omega}\hT\bs{p}^{\times}\bs{Q}(\bs{q}^{-1})\bs{\mathrm{f}} ~~ \text{($\bs{\tau}$ from $\bs{\mathrm{f}}$ in body frame)}\\
		&=\bs{z}\hT\begin{bmatrix}\bs{I}_{3\times3} \\ \bs{p}^{\times}\bs{Q}(\bs{q^{-1}})\end{bmatrix}\bs{\mathrm{f}}\\
		&=\bs{z}\hT\begin{bmatrix}\bs{I}_{3\times3} \\ \bs{p}^{\times}\bs{Q}(\bs{q}^{-1})\end{bmatrix}\begin{bmatrix}
			\bs{b}_1 ~ -\bs{b}_1 ~ \cdots ~ \bs{b}_{\frac{n_\mathrm{f}}{2}} ~ -\bs{b}_{\frac{n_\mathrm{f}}{2}}
		\end{bmatrix}\hT\bs{\beta}\\
		&=\bs{z}\hT\bs{B}\hT\bs{\beta},
	\end{align}
\end{subequations}
where $\bs{\mathrm{f}}$ and $\bs{\tau}$ are the force and torque resulting from the friction at the contact point, $\bs{p}$ is the vector from the center of mass to the contact point in the body frame, and the $^{\times}$ operator creates the skew-symmetric matrix from a vector. The rotation matrix $\bs{Q}(\bs{q}^{-1})$ maps the force $\bs{\mathrm{f}}$ from the global frame to the body frame in order to obtain the torque $\bs{\tau}=\bs{p}^{\times}\bs{Q}(\bs{q}^{-1})\bs{\mathrm{f}}$ in the body frame. The force $\bs{\mathrm{f}}$ at the contact point is the linearized friction force consisting of the basis vectors $\bs{b}_i$ of the friction cone and the magnitude $\bs{\beta}$. The basis vectors $\bno{b}{i}$ of a linearized friction cone are depicted in Fig. \ref{fig:simulator_components} (b).}

\myupdate{The dissipation derivation \eqref{eqn:fric_deriv} can be trivially extended to multiple bodies and contact points. Each contact point involves at most two bodies. Two bodies in contact share the same friction magnitude $\bs{\beta}$ but have different (flipped) basis vectors $\bs{b}_i$. Accordingly, the mapping matrix $\bs{B}$ differs for each body and contact point. Accordingly, the dissipation for multiple bodies and contact points takes on the same form
\begin{equation}
	\frac{\mathrm{d}}{\mathrm{d}t}\mathcal{T}(\dot{\bs{z}}) = \bs{z}\hT\bs{B}\hT\bs{\beta},
\end{equation} 
but now $\dot{\bs{z}}$, $\bs{z}$, and $\bs{\beta}$ are the stacked quantities for all bodies and magnitudes, and $\bs{B}$ consists of the individual $\bs{B}_{i,j}$ matrices of body $i$ and contact point $j$.}

\myupdate{Pairs of basis vectors point in opposite directions. Therefore, all elements of the friction magnitude must be positive. The magnitude is also limited by the friction coefficient $c_\mathrm{f}$ according to Coulomb friction, yielding the constraints
\begin{subequations}
	\begin{align}
		\pmb{1}\hT\bs{\beta} &\leq c_\mathrm{f}\gamma,\\
		\bs{\beta}&\geq \bs{0}.
	\end{align}
\end{subequations}}

\myupdate{Maximum dissipation can, therefore, be stated as a constrained optimization problem
\begin{subequations}
	\begin{alignat}{2}
		&\min_{\bs{\beta}} ~ &&\bs{z}\hT\bs{B}\hT\bs{\beta},\\
		&\text{s.t.}~&&\bs{E}\hT\bs{\beta} \leq \bs{C}_\mathrm{f}\bs{\gamma},\\
		& &&\bs{\beta}\geq \bs{0},
	\end{alignat}
\end{subequations}
which yields the friction forces.}
\newpage

\section{Simulator Algorithms}\label{sec:app_simulator}
The simulator computes the forward dynamics by solving the system of equations \ref{eqn:complete_integrator} at each time step. An interior-point method is implemented as a solver for this system. The implementation follows Algorithm 19.1 in \cite{nocedal_numerical_2006}, which provides more detailed explanations. Pseudo code for the implementation is given in Algorithm \ref{alg:simulator}, and code is available in the open-source implementation (see Section \ref{sec:evaluation}).
\begin{algorithm}[!htb]
	\begin{algorithmic}[1]
		\caption{Dynamics Simulator}\label{alg:simulator}
		\Function{$\mathrm{line\_search}$}{$\Delta\bs{s},\bs{s}^{(0)}$}
			\State $\Delta\bs{s} = \mathrm{feasible\_step}(\Delta\bs{s})$\Comment{remain feasible, see \cite{nocedal_numerical_2006}}
			\State $\alpha=1$ \Comment{step length}
			\While{$\mathrm{true}$}
				\State $\bs{s}^{(1)} = \bs{s}^{(0)} + \alpha\cdot\Delta\bs{s}$\Comment{take a scaled step}
				\If{$\mathrm{norm}(f(\bs{s}^{(1)})) < \mathrm{norm}(f(\bs{s}^{(0)}))$}\Comment{successful}
					\State $\mathrm{return~} \bs{s}^{(1)}$
				\EndIf
				\State $\alpha = \frac{\alpha}{2}$\Comment{half step length}
			\EndWhile
		\EndFunction
		\Function{$\mathrm{interior\_point}$}{$\bs{s}_0$}
			\While{$\mathrm{true}$}\Comment{iterative solver (Newton-based)}
				\State $\mathrm{solve}(\bs{F}(\bs{s}^{(0)})\Delta\bs{s} = -\bs{f}(\bs{s}^{(0)}))$\Comment{use algorithms in Section \ref{sec:numerical_solver}}
				\State $\bs{s}^{(1)} = \mathrm{line\_search}(\Delta\bs{s},\bs{s}^{(0)})$
				\If{$\mathrm{norm}(\bs{f}(\bs{s}^{(1)})) < \mathrm{tolerance}$}\Comment{successful}
					\State $\mathrm{return~} \bs{s}^{(1)}$
				\EndIf
				\State $\mathrm{update\_barrier}()$\Comment{update barrier parameter, see \cite{nocedal_numerical_2006}}
				\State $\bs{s}^{(0)} = \bs{s}^{(1)}$
			\EndWhile
		\EndFunction
		\Function{$\mathrm{simulate}$}{$N,\bs{s}_0$}\Comment{simulate $N$ steps starting at $\bs{s}_0$}
			\For{$k = 0:N-1$}\Comment{for each step solve \eqref{eqn:complete_integrator}}
				\State $\bs{s}_{k+1} = \mathrm{interior\_point}(\bs{s}_k)$\Comment{current state as initial guess}
			\EndFor
		\EndFunction
	\end{algorithmic}
\end{algorithm}
\newpage
\section{Application Details}\label{sec:app_applications}
Details on the application examples and pseudo code are given in this appendix. Code is available in the open-source implementation (see Section \ref{sec:evaluation}). 
\paragraph*{Walking Quadruped}
The legs of the quadruped follow a sinusoidal trajectory with five parameters. The front right and back left leg follow the same trajectory, and the front left and back right leg follow the same trajectory offset by a period of $\pi$. The leg trajectories are tracked with proportional-derivative (PD) controllers. Pseudo code for controlling the gait of the quadrupedal robot is stated in Algorithm \ref{alg:walking}.
\begin{algorithm}[!htb]
	\begin{algorithmic}[1]
		\caption{Leg Controller}\label{alg:walking}
		\Function{$\mathrm{leg\_controller}$}{$t,\mathrm{params}$}\Comment{$\mathrm{params}$ found through learning}
			\State $b, a_1, a_2, d_1, d_2 = \mathrm{params}$
			\State $\theta_{\mathrm{d}1} = a_1\cdot\cos(20\pi\cdot t\cdot b + 0) + d_1$\Comment{thigh joint angle 1}
			\State $\theta_{\mathrm{d}2} = a_1\cdot\cos(20\pi\cdot t\cdot b + \pi) + d_1$\Comment{thigh joint angle 2}
			\State $\theta_{\mathrm{d}3} = a_2\cdot\cos(20\pi\cdot t\cdot b - \frac{\pi}{2}) + d_2$\Comment{calf joint angle 1}
			\State $\theta_{\mathrm{d}4} = a_2\cdot\cos(20\pi\cdot t\cdot b + \frac{\pi}{2}) + d_2$\Comment{calf joint angle 2}

			\For{$\mathrm{front\_right\_leg}$ and $\mathrm{back\_left\_leg}$}
				\State $\tau_{\mathrm{hip}} = 100\cdot(0-\theta_{\mathrm{hip}}) - 5\cdot\dot{\theta}_{\mathrm{hip}}$\Comment{keep hip joint stiff}
				\State $\tau_{\mathrm{thigh}} = 80\cdot(\theta_{\mathrm{d}1}-\theta_{\mathrm{thigh}}) - 4\cdot\dot{\theta}_{\mathrm{thigh}}$\Comment{track desired angle}
				\State $\tau_{\mathrm{calf}} = 60\cdot(\theta_{\mathrm{d}3}-\theta_{\mathrm{calf}}) - 3\cdot\dot{\theta}_{\mathrm{calf}}$\Comment{track desired angle}
			\EndFor
			\For{$\mathrm{front\_left\_leg}$ and $\mathrm{back\_right\_leg}$}
				\State $\tau_{\mathrm{hip}} = 100\cdot(0-\theta_{\mathrm{hip}}) - 5\cdot\dot{\theta}_{\mathrm{hip}}$
				\State $\tau_{\mathrm{thigh}} = 80\cdot(\theta_{\mathrm{d}2}-\theta_{\mathrm{thigh}}) - 4\cdot\dot{\theta}_{\mathrm{thigh}}$
				\State $\tau_{\mathrm{calf}} = 60\cdot(\theta_{\mathrm{d}4}-\theta_{\mathrm{calf}}) - 3\cdot\dot{\theta}_{\mathrm{calf}}$
			\EndFor
		\EndFunction
	\end{algorithmic}
\end{algorithm}

The idea of the sampling-based learning algorithm is to randomly pick the five gait parameters. If progress is made with these parameters, i.e., the robot walks further than before, then the next sampling will be biased in the successful parameter direction. If no progress is made, new parameters are picked randomly in the vicinity of the current parameters. The pseudo-code of the sampling-based learning algorithm for the quadruped is stated in Algorithm \ref{alg:learning}.
\begin{algorithm}[!htb]
	\begin{algorithmic}[1]
		\caption{Sampling-Based Learning Algorithm}\label{alg:learning}
		\State $\mathrm{explore\_factor} = 0.1$\Comment{scaling for random sample generation}
		\State $\mathrm{bias} = \bs{0}_5$\Comment{bias direction for sampling}
		\State $\mathrm{params_0} = \mathrm{params_1} = [0.1, 0, 1, 0, -1.5]$\Comment{initial parameters for standing}
		\State $\mathrm{distance_0} = \mathrm{distance_1} = 0$\Comment{walked distance}
		\For{$\mathrm{episode} = 1:100$}
			\If{$\mathrm{bias} == \bs{0}_5$}\Comment{no improvement in last walk}
				\State $\mathrm{params_1} = \mathrm{params_1} + \mathrm{explore\_factor}\cdot\mathrm{rand}(5)$
			\Else\Comment{improvement in last walk}
				\State $\mathrm{params_1} = \mathrm{params_1} + 0.002\cdot\mathrm{rand}(5)  + 0.01\cdot\mathrm{bias} $
			\EndIf
			\State $\mathrm{distance_1}=\mathrm{simulate}(\mathrm{quadruped},\mathrm{params_1})$\Comment{distance walked in 10s}
			\If{$\mathrm{distance_1} > \mathrm{distance_0}$}\Comment{improvement made}
				\State $\mathrm{bias} = \mathrm{normalize}(\mathrm{params_1} - \mathrm{params_0})$\Comment{calculate bias direction}
				\State $\mathrm{params_0} = \mathrm{params_1}$
				\State $\mathrm{distance_0} = \mathrm{distance_1}$
				\State $\mathrm{explore\_factor} = 0.1$\Comment{reset exploration factor}
			\Else\Comment{no improvement made}
				\State $\mathrm{bias} = \bs{0}_5$
				\State $\mathrm{explore\_factor} = 0.9\cdot\mathrm{explore\_factor}$\Comment{decrease exploration factor}
			\EndIf
		\EndFor
	\end{algorithmic}
\end{algorithm}

\paragraph*{Exoskeleton}
The exoskeleton tracks a sinusoidal rehabilitation trajectory in the shoulder flexion/extension joint and in the elbow flexion/extension joint. A PD-controller with variable scaling is used for tracking, and this scaling is manually tuned with the simulation to adhere to the torque limit. The pseudo-code of the tracking controller is stated in Algorithm \ref{alg:exo}.

\begin{algorithm}[!htb]
	\begin{algorithmic}[1]
		\caption{Exoskeleton Tracking Controller}\label{alg:exo}
		\Function{$\mathrm{exo\_controller}$}{$t,\mathrm{scale}$}\Comment{$\mathrm{scale}$ is tuned manually}
			\State $\theta_{\mathrm{e,d}} = -\frac{\pi}{4} + \frac{\pi}{4}\sin(t)$\Comment{elbow trajectory}
			\State $\theta_{\mathrm{s,d}} = \frac{\pi}{4} + \frac{\pi}{4}\sin(t)$\Comment{shoulder trajectory}
			\State $\tau_{\mathrm{e}} = \mathrm{scale}\cdot 50\cdot(\theta_{\mathrm{e,d}}-\theta_{\mathrm{e}}) + \mathrm{scale}\cdot 5\cdot(\dot{\theta}_{\mathrm{e,d}}-\dot{\theta}_{\mathrm{e}})$\Comment{PD-controller}
			\State $\tau_{\mathrm{s}} = \mathrm{scale}\cdot 100\cdot(\theta_{\mathrm{s,d}}-\theta_{\mathrm{s}}) + \mathrm{scale}\cdot10\cdot(\dot{\theta}_{\mathrm{s,d}}-\dot{\theta}_{\mathrm{s}})$\Comment{PD-controller}
		\EndFunction
	\end{algorithmic}
\end{algorithm}





\end{appendices}


\bibliography{sn-bibliography}


\end{document}